\documentclass[
    preprint,
    5p,
    times,
    twocolumn,
    authoryear,
]{elsarticle}

\usepackage[switch]{lineno}

\usepackage{adjustbox}
\usepackage{placeins}
\usepackage{float}

\usepackage[labelformat=simple, labelfont=bf]{caption}
\usepackage{subcaption} 

\usepackage{xcolor}
\definecolor{gold}{HTML}{F0A91D}  

\usepackage{amssymb,amsmath,amsthm}

\usepackage{booktabs}
\usepackage{makecell}
\usepackage{multirow}
\usepackage{tabularx}
\usepackage{threeparttable}

\usepackage[acronym,nomain,nopostdot,nogroupskip]{glossaries}
\makenoidxglossaries

\usepackage{hyperref}
\hypersetup{
    breaklinks=true,
    colorlinks=true,
    allcolors=blue,
    pdfauthor={Name}
}

\journal{Neural Networks}
\newacronym{au}{AU}{Aleatoric Uncertainty}
\newacronym{bald}{BALD}{Bayesian Active Learning by Disagreement}
\newacronym{be}{BE}{BatchEnsemble}
\newacronym{bvsb}{BvSB}{Best-versus-Second-Best}
\newacronym{bma}{BMA}{Bayesian Model Averaging}
\newacronym{bnn}{BNN}{Bayesian Neural Network}
\newacronym{cde}{CreDE}{Credal Deep Ensemble}
\newacronym{cka}{CKA}{Centered Kernel Alignment}
\newacronym{clue}{CLUE}{Counterfactual Latent Uncertainty Explanation}
\newacronym{elbo}{ELBO}{Evidence Lower Bound Objective}
\newacronym{epce}{EPCE}{Expected Pairwise Cross-Entropy}
\newacronym{eu}{EU}{Epistemic Uncertainty}
\newacronym{epjs}{EPJS}{Expected Pairwise Jensen-Shannon}
\newacronym{epkl}{EPKL}{Expected Pairwise Kullback-Leibler}
\newacronym{ece}{ECE}{Expected Calibration Error}
\newacronym{gp}{GP}{Gaussian Process}
\newacronym{hmc}{HMC}{Hamiltonian Monte Carlo}
\newacronym{js}{JS}{Jensen-Shannon}
\newacronym{kl}{KL}{Kullback-Leibler}
\newacronym{kfac}{KFAC}{Kronecker-factored approximate curvature}
\newacronym{llcm}{LLCM}{Last-Layer Committee Machine}
\newacronym{lle}{LLE}{Last-Layer Ensemble}
\newacronym{lll}{LLL}{Last-Layer Laplace}
\newacronym{map}{MAP}{Maximum a Posteriori}
\newacronym{mod}{MOD}{Maximized Overall Diversity}
\newacronym{msp}{MSP}{Maximum Softmax Probability}
\newacronym{mimo}{MIMO}{Multi-Input Multi-Output}
\newacronym{mhml}{MHML}{Multi-Head Multi-Loss}
\newacronym{mcd}{MCD}{Monte Carlo Dropout}
\newacronym{mcmc}{MCMC}{Markov Chain Monte Carlo}
\newacronym{mi}{MI}{Mutual Information}
\newacronym{mswag}{MultiSWAG}{Multiple Stochastic Weight Averaging Gaussian}
\newacronym{nll}{NLL}{Negative Log-Likelihood}
\newacronym{nngp}{NNGP}{Neural Network Gaussian Process}
\newacronym{ood}{OOD}{Out-of-Distribution}
\newacronym{odin}{ODIN}{Out-of-Distribution Detector}
\newacronym{uq}{UQ}{Uncertainty Quantification}
\newacronym{tu}{TU}{Total Uncertainty}
\newacronym{rbf}{RBF}{Radial Basis Function}
\newacronym{roc}{ROC}{Receiver Operating Characteristic}
\newacronym{se}{SE}{Snapshot Ensemble}
\newacronym{sgld}{SGLD}{Stochastic Gradient Langevin Dynamics}
\newacronym{sgd}{SGD}{Stochastic Gradient Descent}
\newacronym{sngp}{SNGP}{Spectral Normalized Gaussian Process}
\newacronym{swag}{SWAG}{Stochastic Weight Averaging Gaussian}
\newacronym{vbll}{VBLL}{Variational Bayesian Last Layer}
\newacronym{vge}{VGE}{Variance-Gated Ensemble}
\newacronym{vgmu}{VGMU}{Variance-Gated Margin Uncertainty}
\newacronym{vgn}{VGN}{Variance-Gated Normalization}
\newacronym{vi}{VI}{Variational Inference}
\newacronym{llm}{LLM}{Large Language Model}
\newacronym{duq}{DUQ}{Deterministic Uncertainty Quantification}
\newacronym{ddu}{DDU}{Deep Deterministic Uncertainty}
\newacronym{due}{DUE}{Deterministic Uncertainty Estimation}
\newacronym{auroc}{AUROC}{Area Under the ROC Curve}
\newacronym{aupr}{AUPR}{Area Under the Precision-Recall Curve}

\begin{document}
%
\begin{frontmatter}
\title{Uncertainty quantification for trustworthy deep learning: Methods and measures}
\author[inst1]{H. Martin Gillis\corref{cor1}}
\author[inst1]{Thomas Trappenberg}
\cortext[cor1]{Corresponding author. \\
E-mail address: \href{mailto:martin.gillis@.dal.ca}{martin.gillis@.dal.ca} (H. Martin Gillis).}
\affiliation[inst1]{
organization={Faculty of Computer Science, Dalhousie University},
city={Halifax},
state={Nova Scotia},
country={Canada}
}
\begin{abstract}  
The deployment of deep neural networks in safety-critical domains demands reliable estimates of predictive confidence, yet conventional architectures lack principled uncertainty quantification.
This survey provides a structured, critical review of methods for Uncertainty Quantification (UQ) in deep learning, scoped to ensemble-based and approximate Bayesian approaches and the measures used to summarize their outputs.
Relative to existing UQ surveys, our contribution is depth on efficient ensemble approximations and single-pass methods, and a unified treatment that separates the method producing a predictive distribution from the measure that summarizes its uncertainty.
We organize methods into five families: Bayesian neural networks, Monte Carlo Dropout, deep ensembles, efficient ensemble approximations, and last-layer or single-pass approaches.
We situate adjacent work on evidential and prior networks, conformal prediction, and post-hoc calibration, together with the decision-time tasks of out-of-distribution detection and selective prediction.
For each, we examine theoretical motivation, implementation, empirical performance, and limitations.
We then review ensemble diversity theory and uncertainty measures and their decompositions, contrasting the entropy decomposition with pairwise divergence measures, and consolidate evaluation methodology so that our qualitative comparisons share a common basis.
We close with a brief treatment of uncertainty in large language models and open research directions, including efficient epistemic measures for classification, last-layer diversity, diversity and calibration under shift, and hybrid architectures.
\end{abstract}
\end{frontmatter}
\begin{footnotesize}
    \glsfindwidesttoplevelname[\acronymtype]
    \printnoidxglossary[type=\acronymtype,style=alttree,title={Abbreviations},nonumberlist]
\end{footnotesize}
%
\section{Introduction}
\label{sec:introduction}
The deployment of deep neural networks in safety-critical domains such as autonomous driving, medical diagnostics, environmental monitoring, and scientific modeling is now widespread.
However, conventional architectures lack reliable uncertainty estimates accompanying their predictions.
A model that is 95\% confident in an incorrect class provides false assurance and may go unquestioned, whereas a model that reports high uncertainty allows a human to intervene before a decision is made.
Consequently, a substantial body of research addresses \gls{uq} in deep learning, seeking to enable deterministic models with calibrated measures of predictive confidence.
\subsection{Aleatoric and epistemic uncertainty}
It is useful to recall the distinction between aleatoric and epistemic
uncertainty~\citep{Kendall::2017aa,Hullermeier::2021aa}.
Aleatoric uncertainty arises from inherent noise in the data-generating process and is irreducible given the observation model.
Epistemic uncertainty reflects ignorance about the true model parameters and, in principle, is reducible with additional data.
From a Bayesian perspective, epistemic uncertainty is captured by the posterior distribution over model parameters $p(\mathbf{w}\mid\mathcal{D})$.
Methods differ in how they approximate this posterior (or bypass it altogether) and this distinction organizes the sections that follow.
We stress at the outset that the decomposition of predictive entropy into aleatoric/epistemic components is a \emph{modeling convention} rather than a property intrinsic to the data.
The same variability may be labeled aleatoric or epistemic depending on the chosen model class and feature set, a subtlety we return to in \autoref{sec:measures}.
\subsection{Relation to prior surveys}
\label{sec:scope}
\gls{uq} in deep learning is already served by several broad surveys, and a new survey
must justify itself against them.
\citet{Gawlikowski::2023aa} give a comprehensive overview organized primarily by model architecture: Bayesian neural networks, ensembles, single deterministic networks, and test-time augmentation, together with an extensive applications section; their taxonomy is broad but largely orthogonal to the sources of uncertainty a method targets.
\citet{Abdar::2021aa} review \gls{uq} techniques, applications, and challenges across both Bayesian and frequentist methods with a very large bibliography, but provide limited head-to-head comparison of method advantages and disadvantages.
\citet{Hullermeier::2021aa} focus conceptually on the aleatoric and epistemic distinction and its formalization rather than architectures.
\citet{Mena::2021aa} review \gls{uq} for classification primarily from a Bayesian standpoint, under-weighting frequentist and ensemble methods.
Most recently, \citet{He::2025aa} organizes \gls{uq} methods by the uncertainty sources they address, explicitly critiquing architecture-first and Bayesian-first taxonomies.
A recent general review of predictive uncertainty estimation with machine learning offers complementary breadth~\citep{Tyralis::2024aa}, as do a framework-oriented overview~\citep{Zhang::2020aa}, a tutorial pitched at engineering and health-prognostics audiences~\citep{Nemani::2023aa}, and domain-focused reviews such as Bayesian \gls{uq} for image segmentation~\citep{Valiuddin::2025aa}.
\paragraph{What this survey provides}
Rather than attempt broader coverage than the works above, we go deeper on a coherent area of the field and on the measurement question those surveys treat only briefly:
\begin{enumerate}
    \item Method depth where the field has moved fastest: extended treatment of efficient ensemble approximations such as \gls{be}, \gls{se}, \gls{swag}, \gls{mswag}, and \gls{cde}, and last-layer/single-pass approaches including \gls{lll}, \gls{sngp}, \gls{lle}, and deterministic uncertainty methods.
    \item A unified treatment of uncertainty measures. We separate the method that produces a predictive ensemble from the measure that summarizes its uncertainty, and compare the standard mutual-information decomposition against pairwise divergence measures.
    \item An explicit evaluation section (\autoref{sec:eval}) that grounds the qualitative ratings in the comparison tables in a stated evidentiary basis.
\end{enumerate}
\paragraph{Scope and exclusions}
The survey centers on ensemble-based and approximate Bayesian methods for \emph{classification}, with regression treated where the decomposition differs.
Evidential/prior-network methods (\autoref{sec:evidential}), conformal prediction (\autoref{sec:conformal}), and post-hoc calibration (\autoref{sec:calibration}) are summarized and related to the core families, and uncertainty in large language models is treated as an emerging adjacent area (\autoref{sec:llm}).
This treatment is deliberately uneven.
The evidential/credal family is discussed more fully because it bears directly on the aleatoric/epistemic decomposition and the measure question (\autoref{sec:axiomatic}), whereas conformal prediction and post-hoc calibration are included as complementary tools rather than for exhaustive coverage.
\subsection{Organization}
\begin{figure*}[!t]
    \centering
    \includegraphics[width=0.95\textwidth]{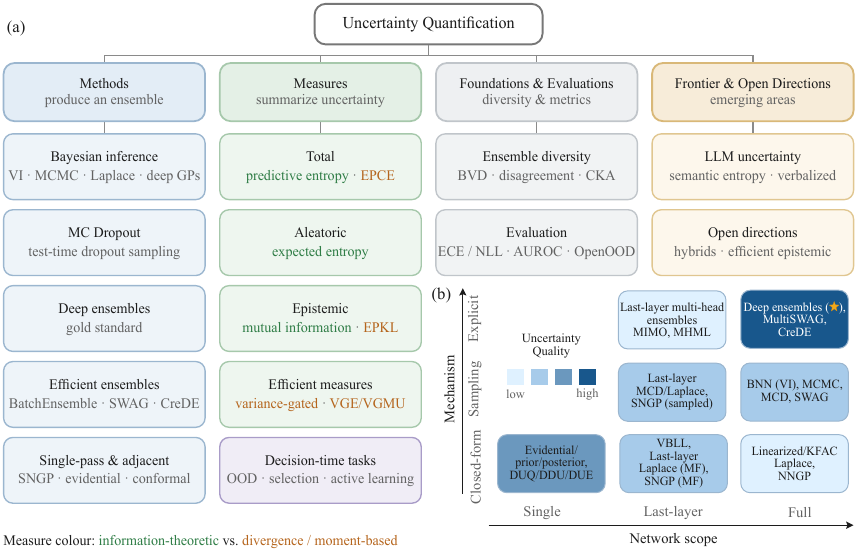}
        \caption{Overview of uncertainty quantification (UQ) in deep learning. 
        \textit{Panel (a)}. Taxonomy used throughout the survey. UQ methods (\hyperref[sec:bnn]{\autoref{sec:bnn}}--\hyperref[sec:method-summary]{\autoref{sec:method-summary}}) produce a predictive ensemble; uncertainty measures (\hyperref[sec:measures]{\autoref{sec:measures}}) summarize that ensemble into total, aleatoric, epistemic, and efficient quantities, and support decision-time tasks such as OOD detection, selective prediction, and active learning (\hyperref[sec:ood]{\autoref{sec:ood}}). 
        Methods and measures are largely decoupled. 
        Among methods that produce a predictive ensemble (explicit members or posterior samples), any such method can be paired with any sample-based measure; closed-form single-pass methods instead expose only their intrinsic measure unless sampled.
        The remaining branches cover foundations and evaluation-ensemble diversity (\hyperref[sec:diversity]{\autoref{sec:diversity}}) and calibration/OOD metrics (\hyperref[sec:eval]{\autoref{sec:eval}}); frontier topics, including LLM uncertainty (\hyperref[sec:llm]{\autoref{sec:llm}}) and open directions (\hyperref[sec:summary]{\autoref{sec:summary}}). 
        Measure labels are colored by approach: information-theoretic (green) \textit{vs}. divergence/moment-based (orange). 
        \textit{Panel (b)}. Method families positioned by network scope (horizontal, increasing left to right: single deterministic network, shared backbone/last layer, full network) and mechanism, how the predictive ensemble is produced (vertical: explicit members, posterior sampling, closed-form predictive). 
        Fill color encodes uncertainty quality, graded from the OOD and calibration ratings of \autoref{tab:methods} (light $=$ near the maximum-softmax-probability baseline, dark $=$ top-tier); deep ensembles are the gold standard (\textcolor{gold}{$\bigstar$}). 
        Quality and computational cost are decoupled: cost rises from left to right and toward the explicit row, so full-network, explicit-ensemble methods are the most expensive, yet high quality also appears at low cost in the bottom-left, where distance-aware single-pass methods (DDU/DUQ/DUE) reach top-tier OOD detection. 
        A closed-form predictive spans all three scopes (single-pass deterministic networks; the last-layer mean-field methods, VBLL and mean-field Laplace/SNGP; and full-network linearized/KFAC Laplace with the infinite-width NNGP limit).
        A single deterministic network admits neither multiple trained members nor a posterior to sample.}
    \label{fig:roadmap}
\end{figure*}

\autoref{sec:framework} introduces the generative-mechanism and network-scope taxonomy that organizes the method families.
\autoref{sec:bnn}--\ref{sec:lastlayer} review the five core method families.
\autoref{sec:evidential}--\autoref{sec:calibration} situate adjacent families.
\autoref{sec:eval} consolidates evaluation methodology (metrics and distribution-shift benchmarks), establishing the evidentiary basis for the comparisons that follow.
\autoref{sec:method-summary} consolidates all method families in comparison tables.
\autoref{sec:diversity} reviews ensemble diversity theory and metrics, and \autoref{sec:measures} treats uncertainty measures and their decompositions, including recent critiques and a sensitivity analysis.
\autoref{sec:ood} treats the decision-time tasks of out-of-distribution detection and selective prediction.
\autoref{sec:llm} orients the reader to the emerging area of uncertainty in large language models.
\autoref{sec:summary} synthesizes the findings (\autoref{tab:methods}--\autoref{tab:measure-class}) and identifies open problems.
\autoref{fig:roadmap} summarizes this structure as a pipeline that deliberately separates the method producing a predictive ensemble from the measure that summarizes its uncertainty.
A chronological overview of the surveyed methods is given in \autoref{tab:timeline-methods}.

\begin{table*}[!t]
\caption{Timeline of uncertainty quantification methods, grouped by family and
ordered by year within each group. References point to the originating work.}
\label{tab:timeline-methods}
\footnotesize
\begin{tabularx}{\textwidth}{@{}l
>{\raggedright\arraybackslash}p{4.0cm}
>{\raggedright\arraybackslash}p{7.5cm}
>{\raggedright\arraybackslash}X@{}}
\toprule
\textbf{Year} & \textbf{Method} & \textbf{Key contribution} & \textbf{Reference} \\
\midrule
\multicolumn{4}{@{}l}{\textit{Bayesian inference}}\\
1992 & Bayesian framework & Practical Bayesian backpropagation & \citet{MacKay::1992aa} \\
1995 & BNN/HMC & Bayesian learning for neural networks & \citet{Neal::1995aa} \\
2011 & Variational inference & Practical VI for networks & \citet{Graves::2011aa} \\
2011 & SGLD & Scalable MCMC via gradient noise & \citet{Welling::2011aa} \\
2015 & Bayes by Backprop & Weight-uncertainty VI & \citet{Blundell::2015aa} \\
2016 & MC Dropout & Dropout as approximate inference & \citet{Gal::2016aa} \\
2018 & NNGP & Infinite-width networks as Gaussian processes & \citet{Lee::2018aa} \\
\addlinespace[6pt]
\multicolumn{4}{@{}l}{\textit{Ensembles}} \\
2015 & TreeNets & Shared-trunk branching ensemble & \citet{Lee::2015aa} \\
2017 & Deep Ensembles & Simple, scalable predictive UQ & \citet{Lakshminarayanan::2017aa} \\
2020 & Anchored Ensembles & Prior-anchored approximate Bayesian ensembling & \citet{Pearce::2020aa} \\
2024 & CreDE & Credal (interval-valued) deep ensembles & \citet{Wang::2024aa} \\
\addlinespace[6pt]
\multicolumn{4}{@{}l}{\textit{Efficient approximations}}\\
2017 & Snapshot Ensembles & Cyclical learning-rate checkpoints as an ensemble & \citet{Huang::2017aa} \\
2019 & SWAG & Gaussian from the SGD trajectory & \citet{Maddox::2019aa} \\
2019 & Sub-Ensembles & Trunk-shared partial ensembles & \citet{Valdenegro-Toro::2019aa} \\
2020 & BatchEnsemble & Rank-one member perturbations & \citet{Wen::2020aa} \\
2020 & MultiSWAG & Multi-basin Bayesian model averaging & \citet{Wilson::2020aa} \\
2021 & Masksembles & Fixed-mask MCD--ensemble interpolation & \citet{Durasov::2021aa} \\
2022 & Layer Ensembles & Single-pass per-layer ensembling & \citet{Kushibar::2022aa} \\
\addlinespace[6pt]
\multicolumn{4}{@{}l}{\textit{Last-layer and single-pass}}\\
2015 & LLE (TreeNets) & Branching at final classification layer & \citet{Lee::2015aa} \\
2020 & Last-layer Bayes & Fixes ReLU overconfidence far from data & \citet{Kristiadi::2020aa} \\
2020 & DUQ & RBF distance to class centroids & \citet{Amersfoort::2020aa} \\
2021 & Laplace redux & Scalable post-hoc Laplace approximation & \citet{Daxberger::2021aa} \\
2021 & MIMO & Multi-input multi-output subnetworks & \citet{Havasi::2021aa} \\
2021 & DUE & Deep-kernel distance-aware UQ & \citet{Amersfoort::2021aa} \\
2022 & SNGP & Distance-aware GP output layer & \citet{Liu::2022aa} \\
2023 & DDU & Feature-density deterministic UQ & \citet{Mukhoti::2023aa} \\
2023 & Multi-Head Multi-Loss & Per-head loss functions & \citet{Galdran::2023aa} \\
2023 & Epinet & Auxiliary epistemic network & \citet{Osband::2023aa} \\
2024 & Repulsive LLE & Function-space repulsion for diversity & \citet{Steger::2024aa} \\
2024 & VBLL & Variational Bayesian last layers & \citet{Harrison::2024aa} \\
\addlinespace[6pt]
\multicolumn{4}{@{}l}{\textit{Evidential, conformal, and calibration}}\\
2017 & Temperature scaling & Post-hoc calibration baseline & \citet{Guo::2017aa} \\
2018 & Evidential deep learning & Dirichlet evidence outputs & \citet{Sensoy::2018aa} \\
2018 & Prior Networks & Dirichlet distributional uncertainty & \citet{Malinin::2018aa} \\
2019 & Conformalized quantile regression & Adaptive distribution-free intervals & \citet{Romano::2019aa} \\
2020 & Posterior Networks & Density-aware Dirichlet, without OOD samples & \citet{Charpentier::2020aa} \\
2023 & Conformal prediction & Finite-sample distribution-free coverage & \citet{Angelopoulos::2023aa} \\
2025 & Flexible evidential deep learning & Flexible-Dirichlet parameterization & \citet{Yoon::2025aa} \\
\bottomrule
\end{tabularx}
\end{table*}

%
\section{Generative mechanisms and network scope}
\label{sec:framework}
We organize the method families that follow along two axes, summarized in \autoref{tab:scope} and visualized in \autoref{fig:roadmap}.
The first axis is the \emph{generative mechanism}, how a method produces the predictive ensemble from which uncertainty is read: by \emph{explicit members} (independently parameterized predictors, deterministic at inference), by \emph{posterior sampling} (Monte Carlo draws from an approximate weight posterior at inference), or by a \emph{closed-form predictive} (an analytic predictive distribution from a single deterministic pass).
The second axis is the \emph{network scope} over which that mechanism operates: the full network, a shared backbone with an independent last layer, or a single deterministic network.
Computational cost generally rises toward explicit members and full-network scope, and the strongest uncertainty (between-mode diversity) arises only in the explicit-member, full-network cell.
We flag the coordinate each family in these axes as it is introduced.
\begin{table*}[!t]
\caption{%
Classification of UQ methods by how the predictive ensemble is generated (mechanism, rows) and network scope (columns, increasing left to right to match \autoref{fig:roadmap}). \textbf{Bold} marks methods that explore multiple loss-landscape modes (between-mode diversity); these consistently give the strongest uncertainty and all occupy the explicit-member, full-network. Computational cost generally rises toward the upper right. A single deterministic network admits neither multiple trained members nor a posterior to sample. Last-Layer Laplace and SNGP each appear under both the sampling and closed-form rows because the same last-layer Gaussian posterior may be consumed either way (by Monte Carlo sampling or by a closed-form mean-field approximation), with the closed-form mode the default in each case. The closed-form, full-network (linearized/KFAC Laplace and the NNGP limit) is listed for taxonomic completeness and treated only briefly (\autoref{sec:bnn}); it is not separately rated in \autoref{tab:methods}.
}
\label{tab:scope}
\footnotesize
\begin{tabularx}{\textwidth}{@{}
>{\raggedright\arraybackslash}p{4.0cm}
>{\raggedright\arraybackslash}p{3.5cm}
>{\raggedright\arraybackslash}X
>{\raggedright\arraybackslash}X@{}
}
\toprule
\textbf{Mechanism} & \textbf{Single deterministic network} & \textbf{Shared backbone/last-layer} & \textbf{Full network} \\
\midrule
Explicit members \newline {\footnotesize(trained, deterministic at inference)} & \makecell[c]{---} & Multi-head/Last-Layer Ensemble (MIMO, MHML, LLCM, repulsive LLE, TreeNet, epinet), Layer Ensemble & \textbf{Deep Ensemble}, \textbf{Anchored Ensemble}, \textbf{CreDE}, \textbf{MultiSWAG}, Snapshot Ensemble, BatchEnsemble, Masksembles, Sub-Ensemble \\
\addlinespace[6pt]
Posterior sampling \newline {\footnotesize(Monte Carlo at inference)} & \makecell[c]{---} & Last-Layer MCD, Last-Layer Laplace (sampled), SNGP (sampled) & BNN (VI), BNN (MCMC), MCD, SWAG \\
\addlinespace[6pt]
Closed-form predictive \newline {\footnotesize(analytic, single-pass)} & Evidential/Prior/Posterior Network, DUQ, DDU, DUE & VBLL, Last-Layer Laplace (mean-field), SNGP (mean-field) & Linearized/KFAC Laplace, NNGP (infinite-width limit) \\
\bottomrule
\end{tabularx}
\end{table*}
%
\section{Bayesian neural networks}
\label{sec:bnn}
\noindent
\textit{Mechanism: Posterior sampling (VI, MCMC) or closed-form predictive (linearized/Laplace, NNGP), full-network scope; last-layer variants in \autoref{sec:lll}.}\par\smallskip
\subsection{Framework and Bayesian model averaging}
The Bayesian treatment of neural networks, first developed by \citet{MacKay::1992aa} and \citet{Neal::1995aa}, provides the most principled framework for \gls{uq} (see \citet{Jospin::2022aa} for a practitioner-oriented tutorial and \citet{Arbel::2023aa} for a critical review of the discussions).
Rather than learning a single point estimate of the weight vector $\mathbf{w}$, a \gls{bnn} places a prior $p(\mathbf{w})$ over the weights; a choice that affects the posterior and remains under-examined~\citep{Fortuin::2022aa,Fortuin::2022ab} and seeks the posterior $p(\mathbf{w}\mid\mathcal{D})$ via Bayes' rule.
Predictions follow by marginalizing over this posterior through the posterior predictive distribution
\begin{equation}
\label{eq:bma}
    p(y\mid\mathbf{x}, \mathcal{D})
    =
    \int_{\mathbf{w}}
    p(y\mid\mathbf{x},\mathbf{w}) \, p(\mathbf{w}\mid\mathcal{D})
    \,\mathrm{d}\mathbf{w}.
\end{equation}
This integral, \gls{bma}, captures epistemic uncertainty.
A posterior with substantial parameter uncertainty results in a correspondingly wide predictive distribution.
In the limit of infinite data, the posterior concentrates and epistemic uncertainty vanishes.
The challenge is that the posterior is intractable for all but the simplest networks, motivating the extensive research on approximate inference.
Beyond standard classification and regression, the same Bayesian treatment extends to structured prediction tasks such as deep survival analysis~\citep{Monod::2025aa}.
\subsection{Variational inference}
\citet{Graves::2011aa} and \citet{Blundell::2015aa} developed practical \gls{vi} schemes for \glspl{bnn}; see \citet{Blei::2017aa} for a general treatment and \citet{Hoffman::2013aa} for black-box and stochastic variants that scale \gls{vi} to large datasets.
\emph{Bayes by Backprop}~\citep{Blundell::2015aa} parameterizes a factorized (mean-field) Gaussian approximate posterior
$q_{\boldsymbol\theta}(\mathbf{w})$
and optimizes
$\boldsymbol\theta$
by minimizing
$D_{\mathrm{KL}}\!\left[
q_{\boldsymbol\theta}(\mathbf{w}) \,\|\, p(\mathbf{w}\mid\mathcal{D})
\right]$,
equivalently maximizing the \gls{elbo}.
At inference, weight samples are drawn from the approximate posterior and predictions are averaged to approximate the \gls{bma} integral.
While principled and compatible with backpropagation, this doubles the number of learnable parameters and adds significant overhead in both training and inference.
\subsection{Markov chain Monte Carlo}
An alternative family draws asymptotically exact posterior samples via \gls{mcmc}. \citet{Neal::1995aa} applied \gls{hmc} to neural networks; \citet{Welling::2011aa} proposed \gls{sgld} to scale \gls{mcmc} to large datasets by injecting calibrated noise into \gls{sgd} updates.
\gls{mcmc} is more faithful to the true posterior than \gls{vi} but is computationally expensive, sensitive to sampling hyperparameters, and yields no closed-form posterior.
Recent full-batch \gls{hmc} studies~\citep{Izmailov::2021aa} provide a gold-standard reference posterior and confirm that inexpensive approximations often differ substantially.

\subsection{Linearized and Laplace approximations}
\noindent
\textit{Mechanism: Closed-form predictive, full-network scope; last-layer variants in \autoref{sec:lll}.}\par\smallskip
A complementary route is the Laplace approximation, which fits a Gaussian at a \gls{map} estimate using (an approximation to) the loss Hessian.
Modern, scalable variants use Kronecker-factored or last-layer Hessians and can be applied post-hoc~\citep{Daxberger::2021aa}.
Since the most practical Laplace variants are last-layer, we treat them in \autoref{sec:lll}.
We note here only that the linearized-Laplace view connects \glspl{bnn} to Gaussian-process inference and underpins several single-pass methods. 
The neural-network/\gls{gp} connection is made explicit by the infinite-width \gls{nngp} limit~\citep{Lee::2018aa}, with deep Gaussian processes a related hierarchical construction~\citep{Damianou::2013aa}.
Applied across the full weight space, these linearized and \acrshort{kfac} Laplace approximations return a closed-form predictive in a single pass, as does the \gls{nngp} limit; together they occupy the closed-form, full-network cell of \autoref{tab:scope}, the full-network analogue of the last-layer mean-field methods of \autoref{sec:lll}.
Hybrid schemes that jointly model structural and parametric uncertainty have also been explored~\citep{Hubin::2019aa}.

\subsection{Practical limitations}
Despite their theoretical correctness, \glspl{bnn} have consistently underperformed deep ensembles in empirical benchmarks~\citep{Gustafsson::2020aa,Ovadia::2019aa}.
\citet{Fort::2020aa} offered an explanation.
Commonly used variational approximations concentrate around a single posterior mode, capturing only local uncertainty.
The true posterior of over-parameterized networks is highly multi-modal, with distinct modes (basins) separated by high-loss barriers.
By constraining the approximate posterior to be unimodal, variational \glspl{bnn} systematically underestimate the epistemic uncertainty arising from multiple qualitatively different solutions.
This ``mode collapse'' is a recurring theme in comparisons of Bayesian and ensemble methods.

\section{Monte Carlo Dropout}
\label{sec:mcd}
\noindent
\textit{Mechanism: Posterior sampling, full-network scope.}\par\smallskip
\subsection{Dropout as approximate inference}
Dropout~\citep{Srivastava::2014aa} was introduced as regularization that stochastically zeroes a fraction of hidden units during training, reducing feature co-adaptation.
\citet{Gal::2016aa} re-interpreted it: retaining dropout at test time and performing stochastic forward passes yields predictions whose empirical mean approximates the predictive mean and whose empirical variance estimates predictive uncertainty.
This corresponds to \gls{vi} with a specific approximate posterior, a product of Bernoulli distributions over binary masks multiplied by point-estimated weight matrices.
The appeal of \gls{mcd} lies in its simplicity.
It requires no change to training beyond standard dropout layers and costs only forward passes at inference.

\subsection{Practical limitations}
The Bernoulli variational family constrains the posterior to a narrow region of weight space around the point estimate.
As \citet{Fort::2020aa} showed, this concentrates around a single mode of the loss landscape, failing to capture multi-modal structure.
Consequently \gls{mcd} tends to underestimate epistemic uncertainty, particularly far from the training distribution.
Additional limitations include sensitivity to the dropout rate (which simultaneously controls regularization strength and uncertainty magnitude) and high correlation among sub-networks.
A line of work addresses these by making the dropout distribution learnable or otherwise more expressive~\citep{Boluki::2020aa,Xie::2021aa}, by adopting alternative variational families such as $\alpha$-divergences~\citep{Li::2017aa}, or by dropping connections rather than units~\citep{Mobiny::2021aa}.
\citet{Ovadia::2019aa} confirmed empirically that \gls{mcd} generally under-performs deep ensembles under dataset shift, especially on \gls{ood} detection. 
\gls{mcd} nonetheless remains a widely used baseline and a lightweight option in resource-constrained settings.

\section{Deep Ensembles}
\label{sec:de}
\noindent
\textit{Mechanism: Explicit members, full-network scope.}\par\smallskip
\subsection{Method and empirical performance}
\citet{Lakshminarayanan::2017aa} proposed deep ensembles as a non-Bayesian approach to predictive \gls{uq}, building on a long tradition of ensemble learning \citep{Dietterich::2000aa}.
The method trains $M$ networks independently from different random initializations, each minimizing the \gls{nll}, and averages their predictions.
For regression, at a single input $\mathbf{x}$ each member $m$ outputs a mean $\mu_m(\mathbf{x})$ and variance $\sigma_m^2(\mathbf{x})$, enabling a decomposition of predictive uncertainty into an aleatoric component (the mean of the individual variances) and an epistemic component (the variance of the individual means).
The spread is taken across members at that fixed input; the explicit law-of-total-variance form is given in \autoref{sec:regression} (\autoref{eq:varianced}).
Despite its simplicity, the method is remarkably effective.
Multiple independent benchmarks~\citep{Gustafsson::2020aa,Ovadia::2019aa} establish deep ensembles as the \emph{de facto} gold standard for \gls{uq} in deep learning, consistently outperforming approximate Bayesian methods on calibration, accuracy, and \gls{ood} detection, although their apparent advantage narrows once individual members are properly calibrated~\citep{Rahaman::2021aa}.

\subsection{Loss-landscape perspective}
\citet{Fort::2020aa} explained this success via the loss landscape.
Networks from different seeds converge to distinct modes separated by high-loss barriers, each member sampling a different basin of attraction.
This multi-modal exploration is precisely what unimodal approximate posteriors fail to achieve, accounting for the systematic performance differences.
\citet{Wilson::2020aa} formalized the view that deep ensembles can be understood as approximate Bayesian model averaging over multiple posterior modes, with their success deriving from marginalizing across distinct solutions.
\citet{Wild::2023aa} subsequently established a more rigorous link between deep ensembles and (variational) Bayesian inference. 
Meanwhile, \citet{Pearce::2020aa} showed that regularizing each member toward an independent draw from the prior (anchored ensembling) makes the ensemble approximate posterior inference.

\subsection{Practical limitations}
The principal limitation is linear cost.
Training $M$ independent networks requires $M\times$ the compute, memory, and storage of a single model.
In practice $M=5$ is common, with diminishing returns beyond that point~\citep{Fort::2020aa}.
Although \citet{Wilson::2020aa} argued for a Bayesian interpretation, the method lacks the formal variational or sampling justification of traditional Bayesian approaches.
For large models where even one training run is expensive, the factor-of-five overhead can be prohibitive, motivating the efficient approximations.
Two cautions modulate the gold-standard narrative: (i) \citet{Abe::2022aa} question whether the full ensemble is necessary, finding much of its benefit recoverable at lower cost, and (ii) \citet{Ashukha::2020aa} show that common in-distribution uncertainty metrics overstate ensemble gains unless test-time calibration is controlled.

\section{Efficient ensemble approximations}
\label{sec:efficient}
\noindent
\textit{Mechanism: Explicit members (SWAG: posterior sampling), full-network scope.}\par\smallskip
\subsection{BatchEnsemble}
\citet{Wen::2020aa} introduced BatchEnsemble, in which each member $m$ modifies a shared weight matrix $\mathbf{W}$ via element-wise multiplication with a member-specific rank-one matrix, $\mathbf{W}_m = \mathbf{W} \odot (\mathbf{r}_m \mathbf{s}_m^\top)$, where $\mathbf{r}_m$ and $\mathbf{s}_m$ are trainable member-specific vectors and $\odot$ denotes the element-wise (Hadamard) product.
The parameter overhead is minimal (two vectors per member per layer), and ensemble inference can be vectorized within a single batch.
However, \citet{Zamyatin::2026aa} recently showed that BatchEnsemble members are near-identical in function space, behaving more like a single model than a true ensemble.
The rank-one perturbations are small relative to the shared weights and are overwhelmed by the shared gradient signal during training, resulting in negligible functional diversity and poor \gls{ood} detection.

\subsection{Snapshot ensembles}
\citet{Huang::2017aa} collect checkpoints along a single training trajectory using a cyclical learning-rate schedule.
Each snapshot captures a different point in weight space, forming an ensemble without independent training runs.
Since all snapshots arise from one trajectory, they typically remain within the same basin of attraction, limiting functional diversity relative to independently trained ensembles.

\subsection{Stochastic weight averaging--Gaussian}
\citet{Maddox::2019aa} proposed \gls{swag}, fitting a Gaussian to the \gls{sgd} trajectory in the final training phase; at inference, weight samples approximate \gls{bma}. \gls{swag} achieves good in-distribution calibration with a single training run, but the posterior is confined to a single mode, limiting \gls{ood} detection.
\citet{Wilson::2020aa} showed that combining \gls{swag} posteriors from multiple independently trained networks, \gls{mswag}, recovers much of the gap with deep ensembles, reinforcing that multi-modal exploration matters for high-quality uncertainty.

\subsection{Credal deep ensembles}
\citet{Wang::2024aa} introduced \glspl{cde}, which extend standard deep ensembles by producing interval-valued outputs rather than point predictions.
Using distributionally robust optimization-inspired training, \glspl{cde} learn credal sets, convex sets of probability distributions that quantify both aleatoric and epistemic uncertainty.
Empirically, \glspl{cde} improve \gls{ood} detection over standard ensembles while maintaining comparable in-distribution performance, offering a promising route to representing epistemic uncertainty via imprecise probabilities~\citep{Walley::1991aa}.
\gls{cde} connects to the credal or evidential viewpoint of \autoref{sec:evidential} and to the axiomatic critique of \autoref{sec:axiomatic}.

\subsection{Further efficient approaches}
Several additional methods trade exactness for cost. 
Deep sub-ensembles share a common trunk and ensemble only a subset of layers~\citep{Valdenegro-Toro::2019aa,Valdenegro-Toro::2023aa}; Masksembles interpolate between \gls{mcd} and deep ensembles using a fixed set of overlapping masks~\citep{Durasov::2021aa}; and layer ensembles treat per-layer weight distributions as a single-pass source of ensemble members~\citep{Oleksiienko::2022aa,Kushibar::2022aa}.
As with the methods above, the recurring challenge is that parameter sharing limits functional diversity. 
\citet{Kirsch::2025aa} sharpens this concern for large models, showing that implicit weight sharing can collapse epistemic uncertainty across nominally independent members, an ``ensemble of ensembles'' that behaves like far fewer effective models.
Complementary directions automate ensemble construction through joint neural-architecture and hyperparameter search~\citep{Egele::2022aa,Herron::2020aa}, amortize members through a shared hypernetwork~\citep{Chauhan::2024aa}, or share a backbone with multiple lightweight heads.

\section{Last-layer and single-pass approaches}
\label{sec:lastlayer}
\noindent
\textit{Mechanism: Spans all three strategies: explicit members, posterior sampling, and closed-form predictive, across shared-backbone/last-layer and single-network scope; flagged per method below.}\par\smallskip
A practically important class of methods seeks ensemble-level uncertainty at (near) single-model cost, either by confining the source of uncertainty to the output layer over a shared backbone or by designing a single deterministic network whose forward pass already encodes uncertainty.
This is motivated by evidence that networks need not be fully stochastic to produce useful uncertainty~\citep{Sharma::2023aa} and by analyses that decouple the representation from the uncertainty head~\citep{Brosse::2020aa}.
This shared cost regime, however, is not a single mechanism: it spans all three of the generative strategies of \autoref{tab:scope}. 
Within it, uncertainty may be produced by \emph{explicit members} (multi-head and last-layer ensembles), by \emph{posterior sampling} restricted to the last layer (last-layer dropout, or a sampled last-layer Laplace or \gls{sngp} posterior), or by a \emph{closed-form predictive} read analytically from a single pass (\glspl{vbll}, the mean-field Laplace and \gls{sngp} approximations, and the deterministic uncertainty methods).
We flag the governing mechanism as each method is introduced.

\subsection{Bayesian and stochastic last-layers}
\label{sec:lll}
\noindent
\textit{Mechanism: Posterior sampling or closed-form predictive, last-layer scope.}\par\smallskip
The Laplace approximation builds a Gaussian posterior centered at the \gls{map} estimate from the loss Hessian.
Applied to the last layer only, it is tractable and can be applied post-hoc to any trained network without retraining~\citep{Daxberger::2021aa}.
\citet{Kristiadi::2020aa} showed that even a minimal Bayesian treatment of the last layer measurably reduces the overconfidence of ReLU networks.
The approach inherits two limitations: (i) it assumes a unimodal Gaussian posterior, and (ii) by restricting uncertainty to the last layer, it ignores epistemic uncertainty in the feature representation itself.
A variational alternative, the \gls{vbll}~\citep{Harrison::2024aa}, trains the last-layer posterior deterministically and marginalizes it analytically, returning a closed-form predictive distribution in a single-pass rather than by sampling; it is thus the closed-form member of this last-layer family (\autoref{tab:scope}).
Restricting dropout to the final layer follows the same logic.
Sampling several passes through a last-layer dropout mask yields a cheap uncertainty estimate that \citet{Brosse::2020aa} found clearly improves on a point-estimate softmax, yet recovers only part of the benefit of full-network \gls{mcd} or a deep ensemble, again because the shared representation is treated deterministically.

\subsection{Spectral-normalized neural Gaussian process}
\noindent
\textit{Mechanism: Closed-form predictive (mean-field) or posterior sampling, last-layer scope.}\par\smallskip
\citet{Liu::2022aa} proposed \gls{sngp}, replacing the final dense layer with a random-feature approximation to a Gaussian process and enforcing spectral normalization throughout the backbone.
Spectral normalization makes the learned representation approximately distance-preserving (inputs far from training data in input space remain far in feature space), so the \gls{gp} output layer can assign appropriate uncertainty.
\gls{sngp} matches the calibration and \gls{ood}-detection quality of a deep ensemble at near single-forward-pass cost, despite being a single deterministic model, making it one of the strongest single-model methods.
The \gls{gp} output layer returns a Gaussian over the logits rather than a point, so the class probabilities are the expected softmax under that Gaussian.
This is precisely the \gls{bma} integral of \autoref{eq:bma} specialized to the approximate last-layer posterior of \gls{sngp}, an intractable integral approximated either by Monte Carlo sampling the output logits or, as \citet{Liu::2022aa} do by default, by a closed-form mean-field approximation that rescales the logits using the predictive variance and avoids sampling at inference.
Its combination of distance-aware representations and a probabilistic output layer is a compelling inductive bias, although it does not capture between-mode epistemic uncertainty.

\subsection{Deterministic uncertainty methods}
\noindent
\textit{Mechanism: Closed-form predictive, single deterministic network.}\par\smallskip
A related line of methods avoids sampling entirely.
\gls{duq}~\citep{Amersfoort::2020aa} uses \gls{rbf} distances to class centroids with a gradient penalty enforcing sensitivity to input changes; \gls{ddu}~\citep{Mukhoti::2023aa} fits a Gaussian mixture in a (spectrally regularized) feature space and separates epistemic (feature-density) from aleatoric (softmax entropy) signals; and \gls{due} extends this line with deep-kernel learning over a distance-aware feature space~\citep{Amersfoort::2021aa}.
Orthonormal certificates~\citep{Tagasovska::2019aa} take a different single-model route, learning a set of diverse functions trained to vanish on the training data so that non-zero responses on new inputs signal epistemic, out-of-distribution uncertainty.
These deterministic uncertainty methods share the reliance on distance-aware representations of \gls{sngp} and similarly target \gls{ood} detection with a single-pass.
They are attractive when latency is critical but, like \gls{sngp}, do not represent multi-modal epistemic uncertainty.
\citet{Postels::2022aa} further caution that the reliability of deterministic epistemic estimates is sensitive to architecture and training choices, influencing their single-pass appeal.

\subsection{Last-layer ensembles and the branching-depth continuum}
\noindent
\textit{Mechanism: Explicit members, last-layer scope.}\par\smallskip
\citet{Lee::2015aa} introduced TreeNets, where an ensemble shares a common trunk and branches at a chosen depth into independent sub-networks.
Branching depth trades parameter efficiency against ensemble diversity. 
Last-layer ensembles are the extreme case of a TreeNet, where branching occurs at the final classification layer.
Members share the entire backbone and differ only in output heads.
Several architectures explore this space: 
\citet{Havasi::2021aa} proposed \gls{mimo}; 
\citet{Galdran::2023aa} proposed \gls{mhml} with per-head calibration; 
\citet{Steger::2024aa} proposed repulsive last-layer ensembles that encourage functional diversity via a repulsive loss; and 
\citet{Osband::2023aa} augment a base network with an auxiliary ``epinet'' that models epistemic uncertainty at low additional cost.
Each navigates the same question: \emph{How much functional diversity is achievable when all members share a feature representation?} 
One practical route combines both diversity sources at the head, pairing last-layer ensemble members with Monte Carlo dropout passes so that diversity is drawn from separately-instantiated heads and stochastic passes together~\citep{Lee::2015aa,Brosse::2020aa,Schweighofer::2023aa,Gillis::2026ab,Gillis::2026aa}.
This approach has a classical antecedent: the Bayesian committee machine \citep{Tresp::2000aa}, where predictions are combined from separately-trained experts within a Bayesian framework, and last-layer ensembles (committee machines) can be read as its modern, shared-backbone realization.

\section{Evidential and prior-network approaches}
\label{sec:evidential}
\noindent
\textit{Mechanism: Closed-form predictive, single deterministic network.}\par\smallskip
A distinct family of methods predicts the parameters of a distribution over predictions in a single forward pass, so uncertainty is produced directly rather than derived post hoc from the predictions.
For classification, evidential deep learning~\citep{Sensoy::2018aa} outputs the parameters of a Dirichlet over the class simplex, interpreting predictions as ``evidence'' and reading epistemic uncertainty from the total evidence mass.
Prior networks~\citep{Malinin::2018aa} similarly parameterize a Dirichlet but explicitly train for distributional uncertainty by exposing the model to \gls{ood} data during training, enabling a clean separation of in-distribution aleatoric uncertainty from distributional (epistemic) uncertainty; a later reverse-\gls{kl} training scheme improved their uncertainty and adversarial robustness~\citep{Malinin::2019aa}.
Posterior networks~\citep{Charpentier::2020aa} replace the \gls{ood}-exposure requirement with a normalizing-flow density over the latent space, resulting in density-aware Dirichlet parameters without auxiliary \gls{ood} data.
More recent variants include flexible-Dirichlet parameterizations for greater expressiveness~\citep{Yoon::2025aa}, post-hoc Dirichlet meta-models that equip a pretrained network with evidential outputs~\citep{Shen::2023aa}, and losses that explicitly widen the in-/out-of-distribution representation~\citep{Nandy::2020aa}.

These methods are appealing for their single-pass cost and explicit uncertainty semantics, and they connect directly to the credal/imprecise-probability viewpoint that runs as a thread through this survey: the credal deep ensembles of \autoref{sec:efficient}, the evidential and prior/posterior networks here, and the axiomatic credal-set foundations of \autoref{sec:axiomatic}.
Their known weaknesses are a sensitivity to the choice of evidential regularizer (the loss term controlling how much the Dirichlet concentrates), difficulty calibrating the evidence scale, and a tendency to be overconfident on far-\gls{ood} inputs.
This far-\gls{ood} overconfidence is a property of the base Dirichlet model, which has no built-in mechanism to recognize inputs far from the training data.
Prior networks remove it by training against auxiliary \gls{ood} data, whereas posterior networks remove it with a latent-space density model that requires no \gls{ood} data at all.
The effectiveness of prior networks therefore relies on how well the chosen auxiliary outliers represent the \gls{ood} encountered at test time, a dataset-dependent assumption that posterior networks avoid.

\section{Conformal prediction and distribution-free UQ}
\label{sec:conformal}
\noindent
\textit{Mechanism: Post-hoc and model-agnostic; produces no predictive ensemble of its own and applies to any base predictor at any scope.}\par\smallskip
Conformal prediction~\citep{Angelopoulos::2023aa} provides finite-sample, distribution-free coverage guarantees.
Given an exchangeable calibration set and a target level $1-\alpha$, it returns prediction sets (classification) or intervals (regression) that contain the true label with probability at least $1-\alpha$, regardless of the underlying model.
Complementary to the methods surveyed here, it applies on top of any base predictor, including an ensemble or \gls{bnn}, and converts a heuristic uncertainty score into a calibrated set.
For regression, conformalized quantile regression~\citep{Romano::2019aa} combines quantile regression~\citep{Koenker::1978aa,Tagasovska::2019aa} with conformal calibration to obtain adaptive intervals with coverage guarantees.
The principal caveats are the exchangeability assumption and the fact that marginal coverage does not imply conditional coverage.
Exchangeability requires that the calibration and test points be statistically interchangeable (their joint distribution is unchanged by reordering), which fails under distribution shift when test inputs are drawn differently from the calibration set.
Marginal coverage means the $1-\alpha$ guarantee holds on average over all inputs, but not necessarily within any specific subgroup; conditional coverage, the stronger property of holding for every input or class, is not guaranteed.

\section{Post-hoc calibration}
\label{sec:calibration}
\noindent
\textit{Mechanism: Post-hoc and model-agnostic; produces no predictive ensemble of its own and rescales the confidence of any base predictor.}\par\smallskip
Calibration (the agreement between predicted confidence and empirical accuracy) is used throughout this survey as an evaluation, so we briefly note the methods that target it directly.
Temperature scaling~\citep{Guo::2017aa} rescales logits by a single learned temperature on a held-out set and is a strong, near-free baseline that preserves accuracy; Platt scaling and isotonic regression are classical alternatives \citep{Platt::1999aa,Zadrozny::2002aa}.
Training-time alternatives target calibration directly.
Focal loss reduces overconfidence and has been analyzed as an implicit calibration regularizer~\citep{Mukhoti::2020aa}, while label smoothing~\citep{Zhang::2021aa} and logit normalization~\citep{Wei::2022aa} similarly temper confidence.
Measuring calibration is itself delicate.
\gls{ece} is sensitive to binning and estimator choice~\citep{Vaicenavicius::2019aa}, and overconfidence is not always the dominant failure mode it is assumed to be~\citep{Wang::2021aa}.
Calibration is also architecture-dependent.
Modern networks are often better calibrated out of the box than the older models that motivated much of this literature~\citep{Minderer::2021aa}.
Crucially, post-hoc calibration improves in-distribution calibration but does not by itself confer epistemic awareness or \gls{ood} robustness.
Calibrated confidence can remain confidently wrong under shift~\citep{Ovadia::2019aa}.
Interactions with ensembling are subtle.
Calibrating members can mitigate accuracy--calibration trade-offs under shift~\citep{Kumar::2022aa}, whereas naively combining ensembles with data augmentation can harm calibration~\citep{Wen::2021aa}.
Calibration methods and \gls{uq} methods are therefore complementary.
The former corrects the scale of confidence, the latter aims to produce confidence that moves appropriately with epistemic state.

\section{Evaluation: Metrics and benchmarks}
\label{sec:eval}
Before comparing the method families head to head, we fix the basis on which they are judged.
The quality of an uncertainty estimate is multifaceted.
A method may be well calibrated yet uninformative, or accurate in-distribution yet unreliable under shift. 
Evaluating \gls{uq} therefore requires complementary metrics together with benchmarks that probe behavior across conditions. 
We review calibration, \gls{ood}, and selective-prediction metrics, the shift and corruption suites used to stress them, and the reporting practices that make comparisons meaningful.

\paragraph{Calibration metrics} \gls{ece} \citep{Naeini::2015aa,Guo::2017aa} bins predictions by confidence and averages the difference between confidence and accuracy; it is widely used but sensitive to bin size.
The Brier score~\citep{Brier::1950aa} and negative log-likelihood are proper scoring rules~\citep{Gneiting::2007aa} that jointly reward calibration and sharpness. 

\paragraph{OOD and selective-prediction metrics} \gls{ood} detection is typically scored by \gls{auroc}\glsadd{roc} and \gls{aupr} using an uncertainty score to separate in- from out-of-distribution inputs (the \gls{msp} baseline~\citep{Hendrycks::2017aa} is the standard reference point).
Standardized suites such as OpenOOD~\citep{Yang::2022aa} and Uncertainty Baselines~\citep{Nado::2022aa} consolidate these protocols, datasets, and reference implementations, permitting reproducible comparison.
Selective prediction is summarized by risk--coverage curves.

\paragraph{Shift and corruption benchmarks} The most informative evaluations stress models under distribution shift.
CIFAR-10-C/CIFAR-100-C and ImageNet-C~\citep{Hendrycks::2019aa} apply parameterized corruptions at increasing severity; the shift suite of \citet{Ovadia::2019aa} tracks calibration and accuracy as a function of severity and is the canonical demonstration that ensembles degrade most gracefully.
For \gls{ood} detection, common pairs include CIFAR-10 \textit{vs.}~SVHN/CIFAR-100 and ImageNet \textit{vs.}~iNaturalist/Places.

\paragraph{Reporting practice} Following \citet{Ovadia::2019aa}, we recommend that \gls{uq} methods be evaluated across multiple metrics, since in-distribution calibration is a weak predictor of behavior under shift.
Where possible, methods should be compared at matched compute, because much of the apparent advantage of ensembles is acquired with $M\times$ training cost.

\section{Comparative summary of methods}
\label{sec:method-summary}
Having surveyed the core and adjacent method families (\autoref{sec:bnn}--\autoref{sec:calibration}) and fixed the evaluation basis (\autoref{sec:eval}), we now consolidate the methods here. 
\autoref{tab:methods} compares methods on computational cost and uncertainty quality, and \autoref{tab:methods-procon} lists their advantages and disadvantages; both are read against the generative-mechanism taxonomy of \autoref{tab:scope} (\autoref{sec:framework}).
The \gls{ood} and calibration entries are qualitative three-level ratings (L/M/H), not reproduced metrics.
Each is anchored to the published benchmarks summarized in \autoref{sec:eval} and defined in note (a) of \autoref{tab:methods} relative to two fixed reference points, the \gls{msp} baseline~\citep{Hendrycks::2017aa} and deep ensembles on the standard \gls{ood} pairs and shift suite of \citet{Ovadia::2019aa}.
The \emph{Evidence} column names the principal source behind each rating.
Where direct \gls{ood}-detection or shift benchmarking was unavailable, the rating is inferred from method structure and identified with a dagger ($\dagger$).
These are therefore relative, evidence-anchored placements rather than precise scores, and should be interpreted against \autoref{sec:eval}.
However, two main qualifications apply.
First, the supporting results are drawn from studies that differ in architecture, dataset, and shift type, and since uncertainty rankings are themselves dataset- and shift-dependent, the ratings are indicative placements rather than directly comparable measurements.
Second, ratings marked with a dagger ($\dagger$) are conservative inferences from method structure rather than direct benchmarks and should not be over-interpreted; where a rating is primarily from the originating work of a method, the standardized suites of \autoref{sec:eval} (OpenOOD~\citep{Yang::2022aa}, Uncertainty Baselines~\citep{Nado::2022aa}) offer an independent point of comparison.
\begin{table*}[!t]
\caption{Comparative summary of UQ methods, grouped by family. Cost in units of
single-model training/inference; $M$ = ensemble size, $S$ = stochastic passes,
SN = spectral normalization, GP = Gaussian process. OOD/Calibration ratings (L/M/H)
are defined in note~(a) and summarize the benchmarks of \autoref{sec:eval}; the
\emph{Evidence} column gives the principal source supporting each rating.}
\label{tab:methods}
\begin{threeparttable}
\footnotesize
\setlength{\tabcolsep}{5pt}
\begin{tabularx}{\textwidth}{@{}
>{\raggedright\arraybackslash}p{3.75cm}
l
>{\raggedright\arraybackslash}
l
>{\raggedright\arraybackslash}p{2.5cm}
c
c
>{\raggedright\arraybackslash}p{3.5cm}
@{}
}
\toprule
\textbf{Method} 
& \textbf{Training} 
& \textbf{Inference} 
& \textbf{Uncertainty} 
& \textbf{OOD}\tnote{a} 
& \textbf{Calibration}\tnote{a} 
& \textbf{Evidence} 
\\
\midrule
\multicolumn{7}{@{}l}{\textit{Bayesian inference}}\\
BNN (VI) & $2\times$ & $S\times$ & AU \& EU & M & M & \citet{Ovadia::2019aa} \\
BNN (MCMC) & high & $S\times$ & AU \& EU & M & M & \citet{Izmailov::2021aa} \\
MC Dropout & $1\times$ & $S\times$ & AU \& EU & M & M & \citet{Ovadia::2019aa} \\
Last-Layer Laplace & post-hoc & $1\times$\,(+Hessian) & EU & L & M & \citet{Kristiadi::2020aa} \\
\addlinespace[6pt]
\multicolumn{7}{@{}l}{\textit{Ensembles and efficient approximations}}\\
Deep Ensemble\tnote{b} & $M\times$ & $M\times$ & AU \& EU & H & H & \citet{Ovadia::2019aa,Gustafsson::2020aa} \\
Anchored Ensemble & $M\times$ & $M\times$ & AU \& EU & M--H & M--H & \citet{Pearce::2020aa}\tnote{$\dagger$} \\
BatchEnsemble & $1\times$ & $1\times$\,(batched) & EU & L & L & \citet{Zamyatin::2026aa} \\
Snapshot Ensemble\tnote{c} & $1\times$ & $M\times$ & EU & M & M & \citet{Huang::2017aa}\tnote{$\dagger$} \\
(Multi)SWAG & ($M\times$) $1\times$ & $S\times$ & EU & M & M & \citet{Wilson::2020aa} \\
Sub-Ensemble & $1\times$+ & $M\times$\,(partial) & EU & M & M & \citet{Valdenegro-Toro::2019aa}\tnote{$\dagger$} \\
Masksembles & $1\times$ & $S\times$ & EU & M & M & \citet{Durasov::2021aa} \\
Layer Ensemble & $1\times$ & $1\times$ & EU & M & M & \citet{Oleksiienko::2022aa}\tnote{$\dagger$} \\
CreDE\tnote{d} & $M\times$ & $M\times$ & AU \& EU & M+ & M+ & \citet{Wang::2024aa} \\
\addlinespace[6pt]
\multicolumn{7}{@{}l}{\textit{Single-pass and last-layer}}\\
SNGP & $1\times$\,(+SN) & $1\times$ & AU \& EU & M--H & M--H & \citet{Liu::2022aa} \\
DUQ/DDU/DUE & $1\times$\,(+regression) & $1\times$ & EU (feature-density) & H & M--H & \citet{Mukhoti::2023aa} \\
TreeNets/LLE & $1\times$ & $1\times$\,(batched) & EU & L--M & M & \citet{Lee::2015aa,Havasi::2021aa}\tnote{$\dagger$} \\
\addlinespace[6pt]
\multicolumn{7}{@{}l}{\textit{Distributional and post-hoc}}\\
Evidential/Prior/Posterior Network & $1\times$ & $1\times$ & AU \& EU (Dirichlet) & M--H\tnote{e} & M & \citet{Charpentier::2020aa} \\
Conformal wrapper & +calibration set & negligible & set/interval coverage & n/a\tnote{f} & guaranteed\tnote{g} & \citet{Angelopoulos::2023aa} \\
Post-hoc calibration & +calibration set & $1\times$ & rescaled confidence & n/a & in-distribution only\tnote{h} & \citet{Guo::2017aa} \\
\bottomrule
\end{tabularx}
\begin{tablenotes}\footnotesize
\item[a] OOD/calibration quality relative to two reference points: \textbf{H} = consistently top-tier, comparable to deep ensembles on standard OOD pairs (CIFAR-10 vs.\ SVHN/CIFAR-100, ImageNet vs.\ iNaturalist/Places) and the shift suite of \citet{Ovadia::2019aa}; \textbf{M} = clearly above the maximum softmax probability baseline~\citep{Hendrycks::2017aa} but below deep ensembles; \textbf{L} = near or below that baseline. Intermediate marks (L--M, M--H, M+) denote placement between adjacent levels.
\item[b] Explores multiple modes of the loss landscape.
\item[c] Limited multi-modal exploration (single basin).
\item[d] Improvement over the standard deep ensemble baseline; rating from the proposing work, with limited independent benchmarking to date.
\item[e] Higher with OOD exposure (prior networks) or density modeling (posterior networks).
\item[f] Conformal targets coverage, not an OOD score \textit{per se}.
\item[g] Marginal coverage guarantee under exchangeability.
\item[h] Improves in-distribution calibration only; no epistemic awareness.
\item[$\dagger$] Rating inferred from method structure; limited direct \gls{ood}-detection or shift benchmarking in the cited work.
\end{tablenotes}
\end{threeparttable}
\end{table*}
\begin{table*}[!t]
\caption{Advantages and disadvantages of uncertainty quantification methods, grouped by family.}
\label{tab:methods-procon}
\footnotesize
\begin{tabularx}{\textwidth}{@{}
>{\raggedright\arraybackslash}p{2.6cm}
>{\raggedright\arraybackslash}X
>{\raggedright\arraybackslash}X@{}
}
\toprule
\textbf{Method} & \textbf{Advantages} & \textbf{Disadvantages} \\
\midrule
\multicolumn{3}{@{}l}{\textit{Bayesian inference}}\\
BNN (VI) & Principled Bayesian framework; captures weight uncertainty; naturally regularizes; compatible with backpropagation. & Doubles parameters; expensive; concentrates on a single mode; underperforms ensembles under shift. \\
BNN (MCMC) & Asymptotically exact posterior samples; gold-standard reference posterior; more faithful than variational approximations. & Very expensive; sensitive to sampling hyperparameters; no closed-form posterior; hard to scale. \\
MC Dropout & Simple; no architecture changes; reuses existing dropout layers; cheap training. & Restricted to a single mode; correlated sub-networks; sensitive to dropout rate; underestimates OOD uncertainty. \\
Last-Layer Laplace & Post-hoc; no retraining; tractable Hessian; closed-form posterior. & Ignores feature uncertainty; over-confident without representation constraints; limited to the last layer. \\
\addlinespace[6pt]
\multicolumn{3}{@{}l}{\textit{Ensembles and efficient approximations}}\\
Deep Ensemble & Multi-modal posterior exploration; approximate BMA over modes; strong calibration and OOD detection; no special architecture. & Linear cost scaling in $M$; high memory footprint; diminishing returns beyond $M=5$. \\
Anchored Ensemble & Prior-anchored regularization makes the ensemble approximate Bayesian posterior inference. & Linear cost in $M$; exact posterior recovery only under restrictive conditions. \\
BatchEnsemble & Near single-model cost; vectorized inference; minimal parameter overhead. & Near-identical members; behaves like a single model; poor OOD detection. \\
Snapshot Ensemble & Ensemble for the cost of one training run; simple cyclical learning-rate schedule. & Limited functional diversity; stays within a single basin; inferior to independent ensembles. \\
(Multi)SWAG & Cheap posterior from the SGD trajectory; closed-form Gaussian; combinable across runs for multi-modal coverage. & Standard SWAG confined to one mode; modified learning-rate schedule; MultiSWAG costs $M\times$ training. \\
Sub-Ensemble & Shares a trunk; cheaper than full ensembles; diversity tunable via branch depth. & Diversity limited by the shared trunk; branch passes still scale with $M$. \\
Masksembles & Interpolates between MC Dropout and deep ensembles using fixed complementary masks; tunable diversity/cost. & Single-basin like MC Dropout at low mask diversity; requires mask-count/overlap tuning. \\
Layer Ensemble & Single-pass uncertainty from per-layer weight distributions; low overhead. & Limited functional diversity (within one trained model). \\
CreDE & Interval-valued (credal) outputs; improved OOD detection over standard deep ensembles. & Same cost as deep ensembles; relatively new with limited cross-domain benchmarking. \\
\addlinespace[6pt]
\multicolumn{3}{@{}l}{\textit{Single-pass and last-layer}}\\
SNGP & Single forward pass; distance-aware; strong calibration; composable with ensembles. & Requires spectral normalization; quality depends on the number of random features; single-mode only. \\
DUQ/DDU/DUE & Single forward pass; distance/density-aware OOD detection; conceptually simple. & Single-mode; sensitive to feature regularization; gradient-penalty/Gaussian-mixture fitting overhead. \\
LLE/LLCM & Near single-model cost; ensemble diversity from output heads; composable with any backbone. & Diversity limited by the shared representation and shared training signal (common batches/ordering); may collapse under shared gradients; achievable diversity is an open question. \\
\addlinespace[6pt]
\multicolumn{3}{@{}l}{\textit{Distributional and post-hoc}}\\
Evidential/Prior/Posterior Network & Single pass; explicit uncertainty semantics; clean aleatoric/epistemic separation (with OOD exposure or density). & Regularizer-sensitive; evidence-scale calibration is hard; far-OOD overconfidence without exposure/density. \\
Conformal & Model-agnostic; finite-sample, distribution-free coverage guarantees; wraps any base predictor. & Exchangeability assumption (violated under shift); marginal, not conditional, coverage. \\
Post-hoc calibration & Near-free; preserves accuracy; strong in-distribution calibration baseline. & Improves in-distribution calibration only; no epistemic awareness; degrades under shift. \\
\bottomrule
\end{tabularx}
\end{table*}


\section{Ensemble diversity}
\label{sec:diversity}
\subsection{Definition and the bias--variance--diversity decomposition}
The comparison in \autoref{sec:method-summary} repeatedly traced uncertainty quality back to how much functional diversity a method achieves; we now make that notion precise.
The theoretical motivation for diversity rests on the bias--variance--diversity decomposition.
For an ensemble of $M$ members each producing a predictive distribution $\mathbf{p}_m$ over classes, \citet{Wood::2023aa} proved an exact decomposition of the expected ensemble cross-entropy loss
\begin{align}
\label{eq:bvd}
    -\mathbb{E}_\mathcal{D}\!\left[\mathbf{y} \cdot \log \bar{\mathbf{p}}\right]
    &= -\frac{1}{M}\sum_{m=1}^{M} \mathbf{y} \cdot \log \mathring{\mathbf{p}}_m
    && \text{(bias)} \nonumber \\
    &\quad + \frac{1}{M}\sum_{m=1}^{M} \mathbb{E}_\mathcal{D}\!\left[D_{\mathrm{KL}}\!\left(\mathring{\mathbf{p}}_m \,\|\, \mathbf{p}_m\right)\right]
    && \text{(variance)} \nonumber \\
    &\quad - \mathbb{E}_\mathcal{D}\!\left[\frac{1}{M}\sum_{m=1}^{M} D_{\mathrm{KL}}\!\left(\bar{\mathbf{p}} \,\|\, \mathbf{p}_m\right)\right]
    && \text{(diversity)},
\end{align}
where $\mathbf{p}_m$ is the $m$-th member predictive distribution and the ensemble prediction $\bar{\mathbf{p}}$ is the centroid combiner of \citet{Wood::2023aa}, here the normalized geometric mean of the member distributions.
The term $\mathring{\mathbf{p}}_m$ is the centroid of member $m$ prediction with respect to the training-set distribution (its expected prediction over random training draws $\mathcal{D}$), the reference against which the bias and variance terms are measured.
The diversity term is the average \gls{kl} divergence from $\bar{\mathbf{p}}$ to each member $\mathbf{p}_m$.
Since this term subtracts from the loss, greater diversity always benefits the ensemble for a given bias and variance.
This generalizes the classical regression decomposition of \citet{Ueda::1996aa}, which partitions expected squared-error loss into bias, variance scaled by $1/M$, and pairwise covariance scaled by $(1-1/M)$~\citep{Brown::2005aa}.
Diversity is thus a formally defined component of ensemble loss.
A subtlety worth emphasizing: the ``diversity always helps'' statement holds for fixed bias and variance; in practice, interventions that increase diversity often change bias and variance simultaneously, so the net effect on loss is not guaranteed.

\subsection{Diversity metrics}
\citet{Kuncheva::2003aa} systematically studied ten diversity measures for classifier ensembles, categorized as pairwise and non-pairwise.
The most widely used are summarized below.

\paragraph{Pairwise disagreement} For classifiers $h_i,h_j$, the disagreement is $D_{ij}=P\big(h_i(\mathbf{x})\neq h_j(\mathbf{x})\big)$, averaged over pairs.
Simple and interpretable, but it treats all disagreements equally regardless of whether they improve the ensemble.

\paragraph{Cohen's $\kappa$} Agreement corrected for chance; values near $1$ indicate low diversity, near $0$ indicate chance-level agreement (high diversity), and negative values indicate worse-than-chance agreement.

\paragraph{Prediction (vote) entropy} A non-pairwise measure
\begin{equation}
\label{eq:voteentropy}
    H = -\frac{1}{M}\sum_{c=1}^{C} n_c \log \frac{n_c}{M},
\end{equation}
where $n_c$ is the number of members predicting class $c$.
It is the hard-vote analogue of the predictive entropy of \autoref{eq:predentropy}; the entropy/mutual-information decomposition of \autoref{sec:measures} formalizes how its epistemic part is separated from the aleatoric.

\paragraph{Variance of predictions} For classification, the variance of predicted probability vectors across members at a given input; for regression, $\mathrm{Var}[p_i(\mathbf{x})]$, which is exactly the epistemic component of the variance decomposition (\autoref{eq:varianced}).

\paragraph{Prediction cosine similarity} The cosine similarity between members' predicted probability (or logit) vectors at a given input; low similarity (equivalently, high cosine distance) indicates that members distribute mass differently and serves as a scale-invariant proxy for functional diversity. \citet{Fort::2020aa} used prediction cosine similarity together with functional distance to show that independently initialized networks occupy distinct loss-landscape modes.

\paragraph{Centered kernel alignment} A representation-level measure, \gls{cka}, comparing internal features across members.
Two networks with low \gls{cka} similarity have learned qualitatively different representations even when their training-set predictions agree, which is relevant for \gls{ood} detection, where representational diversity can yield disagreement on novel inputs.

\paragraph{Vendi score} A recent information-theoretic diversity metric, defined as the exponential of the Shannon entropy of the eigenvalues of a member-similarity matrix, that quantifies effective ensemble size without reference labels~\citep{Friedman::2023aa}.

\subsection{Diversity across UQ methods}
Functional diversity varies sharply across methods and predicts uncertainty quality.
Deep ensembles achieve high diversity because independently initialized networks reach different loss-landscape modes; \citet{Fort::2020aa} quantified this via prediction cosine similarity and functional distance.
\gls{mcd} and \gls{swag} produce members confined to a single basin with high mutual similarity. 
BatchEnsemble members, despite distinct parameterization, are near-identical in function space~\citep{Zamyatin::2026aa}.
Snapshot ensembles achieve intermediate diversity.
For last-layer approaches, diversity can only arise in the output mapping; with a shared backbone, independent random initialization can break symmetry among heads, but how strongly shared gradient pressure converges them remains open.
Strategies for enhancing multi-head diversity \gls{mod}~\citep{Jain::2020aa}, adversarial diversity~\citep{Rame::2021aa}, repulsive function-space inference~\citep{Steger::2024aa}, and per-head weighted losses~\citep{Galdran::2023aa}) remain an active area of research.

\section{Uncertainty measures and decompositions}
\label{sec:measures}
The methods above produce sets of predictions (from ensemble members, stochastic passes, or posterior samples), but quantifying the uncertainty in those predictions is itself non-trivial.
This section reviews the principal measures, emphasizing information-theoretic decompositions and divergence-based measures for classification; regression is treated in \autoref{sec:regression}. 
\autoref{tab:timeline-measures} places these measures on a timeline for an overview.
\begin{table*}[!t]
\caption{Timeline of uncertainty measures for ensemble-based deep learning,
grouped by measure category and ordered by year within each group.}
\label{tab:timeline-measures}
\footnotesize
\begin{tabularx}{\textwidth}{@{}
l
>{\raggedright\arraybackslash}p{4.0cm}
>{\raggedright\arraybackslash}p{8.0cm}
>{\raggedright\arraybackslash}X@{}}
\toprule
\textbf{Year} & \textbf{Measure} & \textbf{Key idea} & \textbf{Reference} \\
\midrule
\multicolumn{4}{@{}l}{\textit{Total uncertainty and decompositions}}\\
1965 & Predictive entropy & Information-theoretic roots; total predictive uncertainty & \citet{Ash::1965aa} \\
2017 & Variance decomposition & Regression aleatoric/epistemic split & \citet{Lakshminarayanan::2017aa} \\
2018 & Entropy/MI decomposition & Additive total/aleatoric/epistemic decomposition & \citet{Depeweg::2018aa,Smith::2018aa} \\
2023 & EPKL/EPCE & Pairwise divergence measures; epistemic (EPKL) and total (EPCE) & \citet{Schweighofer::2023aa} \\
2026 & Variance-gated (VGMU, VGN) & Variance-gated total/aleatoric/epistemic decomposition; epistemic margin score & \citet{Gillis::2026aa} \\
\addlinespace[6pt]
\multicolumn{4}{@{}l}{\textit{Epistemic measures}}\\
2005 & JS-divergence acquisition & JS divergence for active learning & \citet{Melville::2005aa} \\
2011 & BALD & Mutual information & \citet{Houlsby::2011aa} \\
2023 & MI critique & MI underestimates epistemic uncertainty under finite ensembles & \citet{Wimmer::2023aa} \\
2023 & Pairwise-distance estimators & Bound entropy via pairwise divergences & \citet{Berry::2023aa} \\
2023 & Axiomatic epistemic measure & Credal-set desiderata for epistemic uncertainty & \citet{Sale::2023aa} \\
2025 & Integral imprecise metrics & Credal-set imprecise-probability metrics & \citet{Chau::2025aa} \\
\addlinespace[6pt]
\multicolumn{4}{@{}l}{\textit{Aleatoric measures}}\\
2022 & $\beta$-NLL & Training objective for the aleatoric variance head; corrects heteroscedastic NLL & \citet{Seitzer::2022aa} \\
\addlinespace[6pt]
\multicolumn{4}{@{}l}{\textit{Ensemble diversity}}\\
1996 & Bias--variance (regression) & Ensemble error decomposition; diversity as the covariance term & \citet{Ueda::1996aa} \\
2003 & Classifier diversity measures & Ten pairwise/non-pairwise metrics & \citet{Kuncheva::2003aa} \\
2023 & Bias--variance--diversity & Exact ensemble cross-entropy decomposition & \citet{Wood::2023aa} \\
\bottomrule
\end{tabularx}
\end{table*}
%
\subsection{Total predictive entropy}
The most direct measure of total predictive uncertainty for classification is the entropy of the averaged predictive distribution. 
With $M$ members (or $S$ passes), $p(y\mid\mathbf{x},\mathcal{D})=\frac{1}{M}\sum_m p(y\mid\mathbf{x},\mathbf{w}_m)$ (noting that the aggregation rule matters, as averaging logits rather than probabilities shifts the resulting confidence and calibration~\citep{Tassi::2022aa})
\begin{equation}
\label{eq:predentropy}
    H[p(y\mid\mathbf{x},\mathcal{D})] =
    -\sum_{c=1}^{C} p(y=c\mid\mathbf{x})\log p(y=c\mid\mathbf{x}).
\end{equation}
Predictive entropy is high whether the input is inherently ambiguous (aleatoric) or the model lacks knowledge (epistemic); used alone it cannot separate the two.
This is a critical limitation when only the epistemic part is actionable.
In active learning, selecting examples by total entropy directs the labeling effort on irreducibly noisy inputs that also score high in aleatoric uncertainty, whereas the goal is to query inputs the model could still learn from.
The same conflation can make predictive entropy actively misleading as an \gls{ood} score~\citep{Kirsch::2021aa}.

\subsection{The entropy/mutual-information decomposition}
The information-theoretic decomposition~\citep{Ash::1965aa}, formalized by \citet{Depeweg::2018aa} and popularized by \citet{Smith::2018aa}, disentangles the two sources
\begin{align}
\label{eq:decomp}
    \underbrace{H[p(y\mid\mathbf{x},\mathcal{D})]}_{\text{TU}}
    ={}
    & \underbrace{\mathbb{E}_{\mathbf{w}\sim p(\mathbf{w}\mid\mathcal{D})} \big[H[p(y\mid\mathbf{x},\mathbf{w})]\big]}_{\text{AU}} \nonumber \\
    &+ \underbrace{MI[y;\mathbf{w}\mid\mathbf{x},\mathcal{D}]}_{\text{EU}}.
\end{align}
The expected conditional entropy captures \gls{au}; the \gls{mi} between prediction and parameters captures \gls{eu}; together they constitute the \gls{tu}.
In the ensemble approximation
\begin{equation}
\label{eq:miapprox}
    MI[y;\mathbf{w}\mid\mathbf{x},\mathcal{D}] \approx
    H[p(y\mid\mathbf{x},\mathcal{D})]
    - \frac{1}{M}\sum_{m=1}^{M} H[p(y\mid\mathbf{x},\mathbf{w}_m)],
\end{equation}
the difference between mixture entropy and average member entropy. 
This additive decomposition is standard for active learning~\citep{Houlsby::2011aa,Gal::2017aa}, selective prediction~\citep{Geifman::2017aa}, and \gls{ood} detection~\citep{Smith::2018aa}.
Decomposition-aware estimators tailored to ensembles refine the finite-$M$ estimates~\citep{Liu::2019aa}, and supervised variants train directly for the aleatoric/epistemic split~\citep{Li::2025ab}.

\subsection{Critiques of the additive decomposition}
The decomposition has drawn recent criticism for its finite-ensemble behavior. \citet{Wimmer::2023aa} showed \gls{mi} can underestimate epistemic uncertainty in multi-class settings.
When members disagree but each is confident in a different class, average entropy (\gls{au}) is low while mixture entropy (\gls{tu}) is high, giving high \gls{mi}; but when members produce diffuse but distinct distributions, both are high, so \gls{mi} stays low despite genuine disagreement.
The decomposition also requires access to individual member predictions.
These issues motivated \citet{Schweighofer::2023aa} to propose pairwise divergence measures that bypass the standard entropy decomposition.

\subsection{Pairwise divergence measures}
\label{sec:pairwise}
\citet{Schweighofer::2023aa} argued that the standard additive decomposition implicitly assumes the \gls{bma} distribution equals the true posterior predictive (an assumption that breaks under finite ensembles) and proposed pairwise measures that avoid it.
For total predictive uncertainty they introduced the \gls{epce}
\begin{equation}
\label{eq:epce}
    \text{EPCE} = \frac{1}{M^2}\sum_{i=1}^{M}\sum_{j=1}^{M}
    H\big[p(y\mid\mathbf{x},\mathbf{w}_i),\, p(y\mid\mathbf{x},\mathbf{w}_j)\big],
\end{equation}
with $H[p,q]=-\sum_c p(c)\log q(c)$. 
For epistemic uncertainty they introduced the \gls{epkl} divergence,
\begin{equation}
\label{eq:epkl}
    \text{EPKL} = \frac{1}{M^2}\sum_{i=1}^{M}\sum_{j=1}^{M}
    D_{\mathrm{KL}}\big[p(y\mid\mathbf{x},\mathbf{w}_i)\,\|\,p(y\mid\mathbf{x},\mathbf{w}_j)\big],
\end{equation}
which decomposes as the sum of \gls{mi} and the reverse \gls{mi} and upper-bounds \gls{mi} by Jensen's inequality~\citep{Schweighofer::2023aa}; the pairwise-\gls{kl} form also appears in \citet{Malinin::2018aa}.
\gls{epkl} is zero when all members agree and grows with disagreement.
Its weaknesses are that \gls{kl} is asymmetric, $D_{\mathrm{KL}}[p_i\|p_j]\neq D_{\mathrm{KL}}[p_j\|p_i]$, and unbounded, causing instability when a member assigns near-zero probability to a class.

A remedy replaces \gls{kl} with the symmetric, bounded Jensen--Shannon divergence
\begin{equation}
\label{eq:js}
    D_{\mathrm{JS}}[p_i\|p_j]=\tfrac{1}{2}D_{\mathrm{KL}}[p_i\|m]
    +\tfrac{1}{2}D_{\mathrm{KL}}[p_j\|m],~m=\tfrac{1}{2}(p_i+p_j)
\end{equation}
bounded in $[0,1]$ (using base-2 logarithms throughout this discussion), giving the \gls{epjs} divergence
\begin{equation}
\label{eq:epjs}
    \text{EPJS} = \frac{1}{M^2}\sum_{i=1}^{M}\sum_{j=1}^{M}
    D_{\mathrm{JS}}\big[p(y\mid\mathbf{x},\mathbf{w}_i)\,\|\,p(y\mid\mathbf{x},\mathbf{w}_j)\big].
\end{equation}

\noindent
Pairwise-distance epistemic estimators (including \gls{js}-divergence variants) have appeared in the literature.
\citet{Berry::2023aa} develop pairwise-distance estimators that bound entropy via pairwise divergences for regression ensembles and connect them to \gls{bald}, building on a longer line of ensemble-disagreement estimators that quantify epistemic uncertainty through divergences between member predictive distributions, such as the expected pairwise \gls{kl} divergence of \citet{Malinin::2018aa, Malinin::2020aa}. 
The \gls{js} divergence in particular has a long history as an uncertainty signal, \textit{e.g.} for active learning~\citep{Melville::2005aa}, and continues to be used as an epistemic (variance) proxy and full-distribution disagreement score in recent work~\citep{Schirmer::2023aa}.
We note, however, that information-theoretic decompositions of this kind have been shown to behave inconsistently as epistemic measures under posterior mismatch and finite ensembles~\citep{Wimmer::2023aa}.

\subsection{Comparative stability of divergence-based measures}
\label{sec:stability}
The three epistemic measures above (\gls{mi}, \gls{epkl}, \gls{epjs}) differ in structural properties that reflect directly on their numerical behavior (\autoref{tab:divergence}).
\autoref{tab:divergence} describes the symmetry and boundedness of just these three epistemic divergence measures.
The broader description of all uncertainty measures and their classification by type and computational approach is consolidated separately in \autoref{tab:measures} and \autoref{tab:measure-class} (\autoref{sec:measure-summary}).
\begin{table*}[!t]
\caption{Structural properties of the three epistemic measures compared in \autoref{sec:stability}.
Symmetry and boundedness are the properties that govern numerical stability; sensitivity trade-offs against them.}
\label{tab:divergence}
\footnotesize
\begin{tabularx}{\textwidth}{@{}
>{\raggedright\arraybackslash}p{1.8cm}
>{\raggedright\arraybackslash}p{2.6cm}
>{\raggedright\arraybackslash}p{1.8cm}
>{\raggedright\arraybackslash}p{2.4cm}
>{\raggedright\arraybackslash}X@{}
}
\toprule
\textbf{Measure} & \textbf{Basis} & \textbf{Symmetric} & \textbf{Bounded} & \textbf{Principal limitation} \\
\midrule
MI   & Mixture-based      & n/a               & Yes ($\le \log_2 C$ bits)   & Underestimates EU for diffuse-but-distinct members~\citep{Wimmer::2023aa} \\
EPKL & Pairwise (KL)      & No                & No                   & Unbounded; unstable near zero-probability classes \\
EPJS & Pairwise (JS)      & Yes               & Yes ($[0,1]$ bit)   & Less sensitive than EPKL; $\mathcal{O}(M^2C)$ cost \\
\bottomrule
\end{tabularx}
\end{table*}

\gls{epkl} is asymmetric and unbounded.
When member $p_i$ assigns near-zero mass to a class for which $p_j$ assigns non-negligible mass, $D_{\mathrm{KL}}[p_i\|p_j]$ diverges, disproportionately amplifying the aggregate.
This makes \gls{epkl} the most sensitive of the three to distributional differences but also the most fragile, since a single tail-class asymmetry can dominate the score and compress the in-/out-of-distribution margin that governs detection.
This effect is exacerbated under shift, where members lose a shared reference and extrapolate independently.
\gls{epjs} replaces \gls{kl} with the symmetric Jensen--Shannon divergence, bounded in $[0,1]$, so any single class pair's contribution is restricted and the near-zero-probability pathology is removed.
\gls{mi} sidesteps unboundedness altogether by operating on the mixture rather than on pairs, but this is known to underestimate epistemic uncertainty for diffuse-but-distinct members~\citep{Wimmer::2023aa}.

\paragraph{Interpretation:~Sensitivity and stability trade-off} The unboundedness of \gls{epkl} is precisely what makes it both most responsive to member disagreement and most prone to instability, whereas \gls{epjs} and \gls{mi} sacrifice some responsiveness for numerical stability.
The defensible, model-independent claim is structural.
The symmetry and boundedness of \gls{epjs} remove a known failure mode of \gls{epkl}, while its pairwise form avoids the mixture behavior that limits \gls{mi}.
We do not claim that \gls{epjs} is the best \gls{ood} detector; which measure performs best in practice depends on the ensemble and the nature of the shift, and must be established empirically on standard suites (\autoref{sec:eval}).
We present this discussion and comparison as a synthesis of the properties of the measures rather than as a new empirical result.
\subsection{Axiomatic perspectives}
\label{sec:axiomatic}
Beyond empirical comparison, an axiomatic line asks what a proper epistemic measure should satisfy.
\citet{Sale::2023aa}, drawing on proper scoring rules and the geometry of credal sets, argued that a valid epistemic measure should be zero when the posterior concentrates on a single model and increase monotonically with posterior spread, establishing formal desiderata that existing measures meet to varying degrees. 
This links the measure question to the credal/evidential methods of \autoref{sec:efficient} and \autoref{sec:evidential}.
Related formalizations include integral imprecise-probability metrics over credal sets~\citep{Chau::2025aa}, a rate--distortion view of uncertainty quantification~\citep{Apostolopoulou::2024aa}, and epistemic estimation via explicitly adversarial models~\citep{Schweighofer::2023ab}.

\subsection{Efficient variance-gated measures}
\label{sec:vge}
A practical drawback of the pairwise measures above is their $\mathcal{O}(M^2C)$ cost, which becomes prohibitive for large ensembles and many classes. 
To reduce this cost, \glspl{vge}~\citep{Gillis::2025ac,Gillis::2026aa} gate member probabilities by a signal-to-noise factor derived from the ensemble mean and per-class predictive spread, yielding a total/aleatoric/epistemic decomposition at $\mathcal{O}(MC)$, a margin-based epistemic score, \gls{vgmu} extends the \gls{bvsb} criterion~\citep{Joshi::2009aa} at $\mathcal{O}(C)$, and a differentiable training-time calibration layer, \gls{vgn}.
Like the other efficient approximations surveyed here, variance-gating trades the explicit information-theoretic semantics of the entropy decomposition for lower cost, and its behavior under distribution shift has not yet been benchmarked against \gls{mi} and the pairwise measures on the standard suites of \autoref{sec:eval}. 
It nonetheless illustrates the broader search for efficient, non-entropy-based epistemic measures motivated by the critiques of \citet{Wimmer::2023aa}.

\subsection{Variance-based measures for regression}
\label{sec:regression}
For regression the decomposition takes an algebraic form via the law of total variance.
When each member outputs a mean $\mu_m(\mathbf{x})$ and variance $\sigma_m^2(\mathbf{x})$~\citep{Lakshminarayanan::2017aa}
\begin{equation}
\label{eq:varianced}
    \underbrace{\mathrm{Var}[y\mid\mathbf{x}]}_{\text{TU}} =
    \underbrace{\frac{1}{M}\sum_{m=1}^{M}\sigma_m^2(\mathbf{x})}_{\text{AU}}
    + \underbrace{\frac{1}{M}\sum_{m=1}^{M}\big(\mu_m(\mathbf{x})-\bar\mu(\mathbf{x})\big)^2}_{\text{EU}}.
\end{equation}
The mean of individual variances captures aleatoric uncertainty; the variance of individual means captures epistemic uncertainty.
This is exact under a mixture-of-Gaussians assumption. 
For distribution-free intervals, quantile regression~\citep{Koenker::1978aa,Tagasovska::2019aa} and its conformalized variant~\citep{Romano::2019aa} offer alternatives.
Training the aleatoric head is itself delicate.
The Gaussian negative-log-likelihood objective scales the gradient of each point by its predicted variance, systematically under-sampling poorly-fit regions, which $\beta$-weighted variants correct~\citep{Seitzer::2022aa}.
Hybrid models that pair a flexible (\textit{e.g.} normalizing-flow) aleatoric density with a separate epistemic estimator further decouple the two components~\citep{Van-Katwyk::2025aa}.

\subsection{Summary of uncertainty measures}
\label{sec:measure-summary}
Whereas \autoref{tab:divergence} focused only on the three epistemic divergence measures, the two tables here cover the full set: \autoref{tab:measures} compares all the uncertainty measures introduced above on task, cost, strengths, and limitations, and \autoref{tab:measure-class} classifies them by uncertainty type (total, aleatoric, epistemic) and computational approach (information-theoretic \textit{vs.} moment/divergence-based).
\begin{table*}[!t]
\caption{Comparative summary of uncertainty measures, grouped by the uncertainty
they capture. All measures are for classification except where noted. The
\emph{Key reference} column gives the introducing or defining reference for each measure.}
\label{tab:measures}
\begin{threeparttable}
\footnotesize
\begin{tabularx}{\textwidth}{@{}
>{\raggedright\arraybackslash}p{3.5cm}
l
>{\raggedright\arraybackslash}p{4.5cm}
>{\raggedright\arraybackslash}p{4.5cm}
>{\raggedright\arraybackslash}X
@{}
}
\toprule
\textbf{Measure}
& \textbf{Cost}\tnote{a}
& \textbf{Strengths}
& \textbf{Limitations}
& \textbf{Key reference}
\\
\midrule
\multicolumn{5}{@{}l}{\textit{Total uncertainty}}\\
Predictive Entropy & $\mathcal{O}(C)$ & Single scalar; info-theoretic & Does not separate aleatoric/epistemic & \citet{Ash::1965aa} \\
EPCE & $\mathcal{O}(M^2C)$ & Pairwise total uncertainty; avoids BMA-as-posterior assumption & Quadratic in $M$ & \citet{Schweighofer::2023aa} \\
\addlinespace[6pt]
\multicolumn{5}{@{}l}{\textit{Aleatoric uncertainty}}\\
Expected entropy & $\mathcal{O}(MC)$ & Mean of member entropies; aleatoric component of the decomposition & Conflated with total uncertainty if used alone & \citet{Depeweg::2018aa} \\
\addlinespace[6pt]
\multicolumn{5}{@{}l}{\textit{Epistemic uncertainty}}\\
MI & $\mathcal{O}(MC)$ & Principled; expected info gain & Assumes BMA $\equiv$ true posterior; underestimates under finite ensembles & \citet{Houlsby::2011aa} \\
EPKL & $\mathcal{O}(M^2C)$ & Improved epistemic sensitivity; robust to posterior mismatch & Quadratic in $M$; unbounded & \citet{Schweighofer::2023aa} \\
EPJS & $\mathcal{O}(M^2C)$ & Bounded; symmetric; numerically stable & Quadratic in $M$; less sensitive than EPKL & \citet{Schweighofer::2023aa}; \citet{Melville::2005aa} \\
VGMU & $\mathcal{O}(C)$ & Margin-based; decision-focused; epistemic-aware & Uses top-2 classes; insensitive to tail-classes & \citet{Gillis::2026aa} \\
\addlinespace[6pt]
\multicolumn{5}{@{}l}{\textit{Full decompositions (multiple types)}}\\
Variance-gated decomposition & $\mathcal{O}(MC)$ & Linear-time gated decomposition (TU/AU/EU); differentiable & Behavior gating-parameter dependent & \citet{Gillis::2026aa} \\
Variance decomposition\tnote{b} & $\mathcal{O}(M)$ & Exact TU/AU/EU split under Gaussian mixture; tractable & Gaussian assumption & \citet{Lakshminarayanan::2017aa} \\
\bottomrule
\end{tabularx}
\begin{tablenotes}\footnotesize
\item[a] Costs denote the additional cost of computing each measure given the $M$ member predictions and their aggregate $\bar p$; $C$ is the number of classes.
\item[b] Defined for regression; all other measures listed are for classification.
\end{tablenotes}
\end{threeparttable}
\end{table*}
\begin{table*}[!t]
\caption{Classification of uncertainty measures by type and computational approach.}
\label{tab:measure-class}
\footnotesize
\begin{tabularx}{\textwidth}{@{}
>{\raggedright\arraybackslash}p{2.5cm} 
>{\raggedright\arraybackslash}p{5.0cm}
>{\raggedright\arraybackslash}X@{}
}
\toprule
\textbf{Uncertainty} & \textbf{Information-theoretic} & \textbf{Moment/divergence-based} \\
\midrule
Total & Predictive entropy $H[\bar p]$ & EPCE; total predictive variance (regression); variance-gated decomposition \\
Aleatoric & Expected entropy $\mathbb{E}[H[p_m]]$ & Mean variance $\tfrac{1}{M}\sum\sigma_m^2$ (regression); variance-gated decomposition \\
Epistemic & Mutual information $I(y;\mathbf{w}\mid\mathbf{x},\mathcal{D})$ & EPKL, EPJS, VGMU, variance of means (regression); variance-gated decomposition \\
\bottomrule
\end{tabularx}
\end{table*}

\section{Out-of-distribution detection and selective prediction}
\label{sec:ood}
With the methods that produce predictive ensembles (\autoref{sec:bnn}--\autoref{sec:method-summary}) and the measures that summarize their uncertainty (\autoref{sec:measures}) in place, we turn to the two decision-time tasks that ultimately judge them: flagging inputs that fall outside the training distribution, and abstaining when a prediction cannot be trusted.
Both reduce to thresholding an uncertainty score, yet how that score is constructed is itself a substantial research question; one that is largely independent of which method produced the predictive distribution (\autoref{tab:ood}).
A third, closely related decision-time use of the same uncertainty scores is active learning, where high epistemic uncertainty selects the most informative points to label~\citep{Houlsby::2011aa,Gal::2017aa} via the \gls{bald} criterion of \autoref{sec:measures}.
\begin{table*}[!t]
\caption{Out-of-distribution detection and selective-prediction approaches
(\autoref{sec:ood}). Post-hoc scores operate on a fixed trained classifier; training-time and
Bayesian/ensemble approaches modify training or inference. Standardized evaluation: OpenOOD~\citep{Yang::2022aa}; strong pretrained representations raise the achievable ceiling~\citep{Fort::2021aa}.}
\label{tab:ood}
\begin{threeparttable}
\footnotesize
\begin{tabularx}{\textwidth}{@{}
>{\raggedright\arraybackslash}p{4.0cm}
>{\raggedright\arraybackslash}p{6.0cm}
>{\raggedright\arraybackslash}X
>{\raggedright\arraybackslash}X@{}
}
\toprule
\textbf{Approach} 
& \textbf{Signal/mechanism} 
& \textbf{Requirements} 
& \textbf{Representative work} 
\\
\midrule
\multicolumn{4}{@{}l}{\textit{Post-hoc scores (training-free; operate on a fixed model)}}\\
MSP & Maximum class probability (baseline) & None & \citet{Hendrycks::2017aa} \\
ODIN & Temperature scaling + input perturbation & Tune $T,\epsilon$ & \citet{Liang::2020aa} \\
Mahalanobis & Class-conditional feature distance & Fit Gaussians & \citet{Lee::2018ac} \\
Relative Mahalanobis & Near-OOD-corrected distance & Fit Gaussians & \citet{Ren::2021aa} \\
Energy & Free energy of the logits & None & \citet{Liu::2020ab} \\
LogitGap & Gap between top and remaining logits & None & \citet{Liang::2025aa} \\
\addlinespace[6pt]
\multicolumn{4}{@{}l}{\textit{Training-time}}\\
Learned confidence & Auxiliary confidence head & Retraining & \citet{DeVries::2018aa} \\
Outlier exposure & Train against auxiliary outliers & Auxiliary OOD data & \citet{Lee::2018ab} \\
OOD pseudo-inputs & Train against synthesized OOD inputs & Generate inputs & \citet{Segonne::2022aa} \\
Orthonormal certificates & Diverse functions trained to vanish on in-distribution data; non-zero response flags OOD & Train certificates (no OOD data) & \citet{Tagasovska::2019aa} \\
\addlinespace[6pt]
\multicolumn{4}{@{}l}{\textit{Bayesian/ensemble}}\\
BNN/dropout posterior & Posterior disagreement (epistemic) & Sampling & \citet{Nguyen::2022aa} \\
Bayesian + outlier exposure & Posterior with outlier exposure & Auxiliary OOD data & \citet{Wang::2021ab} \\
\addlinespace[6pt]
\multicolumn{4}{@{}l}{\textit{Selective prediction/abstention}}\\
Softmax response & Confidence threshold; risk--coverage & None & \citet{Geifman::2017aa} \\
Trust score & Agreement with $k$-NN class structure & Reference set & \citet{Jiang::2018aa} \\
Calibrated selective classification & Coverage-guaranteed abstention & Calibrated set & \citet{Fisch::2022aa} \\
Reject option (survey) & Taxonomy of abstaining classifiers & None & \citet{Hendrickx::2024aa} \\
\bottomrule
\end{tabularx}
\end{threeparttable}
\end{table*}

\subsection{Score-based and post-hoc detection}
A series of post-hoc refinements improve separability over the \gls{msp} baseline (\autoref{sec:eval}) without retraining: (i) temperature scaling with input perturbation, \gls{odin}~\citep{Liang::2020aa}; (ii) Mahalanobis distance to class-conditional Gaussians in feature space~\citep{Lee::2018ac}, with a relative variant for near-\gls{ood}~\citep{Ren::2021aa}; (iii) the free energy of the logits~\citep{Liu::2020ab}; and, more recently, (iv) scores derived from the gap between the top logit and the remainder~\citep{Liang::2025aa}.
Training-time approaches instead learn an explicit confidence or detection signal, via a confidence branch~\citep{DeVries::2018aa} or by exposing the model to auxiliary outliers~\citep{Lee::2018ab}.
\citet{Fort::2021aa} show that strong pretrained representations can dramatically improve on what these scores can achieve, particularly for near-\gls{ood}.

\subsection{Bayesian and ensemble approaches}
Since \gls{ood} inputs are precisely where epistemic uncertainty should be highest, the method families of this survey are themselves \gls{ood} detectors. 
Bayesian neural networks, dropout-based posteriors~\citep{Nguyen::2022aa}, and ensembles supply epistemic scores directly; combining a Bayesian treatment with outlier exposure~\citep{Wang::2021ab} or with training on synthesized \gls{ood} pseudo-inputs~\citep{Segonne::2022aa} further sharpens detection. 
The standardized suites of \autoref{sec:eval} consolidate these methods and protocols.
A recurring caution (\autoref{sec:measures}) is that predictive entropy alone can be a poor \gls{ood} score~\citep{Kirsch::2021aa}.

\subsection{Selective prediction and abstention}
Selective prediction equips a model with a reject option, trading coverage for accuracy along a risk--coverage curve~\citep{Geifman::2017aa,Hendrickx::2024aa}. 
Practical instantiations range from trust scores that flag likely misclassifications~\citep{Jiang::2018aa} to calibrated selective classifiers with coverage guarantees~\citep{Fisch::2022aa}. 
A complementary response to high uncertainty is to abstain.
The quality of the abstention decision inherits directly from the quality of the underlying uncertainty estimate, relating this task back to the method and measure choices of the preceding sections.
A related decision-time capability is to \emph{explain} the uncertainty, identifying the input features responsible for it via counterfactual approaches such as \gls{clue}~\citep{Antoran::2021aa}; we return to this largely open direction in \autoref{sec:summary}.

\section{Uncertainty in large language models}
\label{sec:llm}
The methods surveyed so far target a categorical (or real-valued) predictive distribution over a fixed output space. 
Autoregressive \glspl{llm} break both assumptions; a prediction is a variable-length token sequence, and many forms express the same meaning, so a token-level probability conflates linguistic variation with genuine semantic uncertainty. 
This has caused a fast-growing and still largely heuristic research literature, which we present briefly here (\autoref{tab:llm}) and which two recent surveys cover in depth~\citep{Shorinwa::2024aa,Huang::2024ab}. 
The area is an emerging adjacent field rather than the main focus of the present survey.
\begin{table*}[!t]
\caption{Families of uncertainty quantification for large language models. \emph{Access} indicates whether a method needs model internals/logits (white-box), only sampled output text (black-box), or a held-out calibration set. \emph{Cost} is in forward passes/samples per query; $K$ = number of sampled generations, $M$ = ensemble size.}
\label{tab:llm}
\begin{threeparttable}
\footnotesize
\begin{tabularx}{\textwidth}{@{}
>{\raggedright\arraybackslash}p{2.7cm}
l
>{\raggedright\arraybackslash}X
l
>{\raggedright\arraybackslash}X@{}
}
\toprule
\textbf{Family} & \textbf{Access}\tnote{a} & \textbf{Uncertainty signal} & \textbf{Cost} & \textbf{Representative work} \\
\midrule
Token likelihood & White & Sequence log-likelihood/per-token entropy (length-biased) & $1$ & \citet{Malinin::2021aa} \\
Semantic entropy & White\tnote{b} & Entropy over meaning-equivalence clusters of sampled generations & $K$ ($+$entailment) & \citet{Kuhn::2023aa,Farquhar::2024aa} \\
Sampled consistency & Black & Agreement/similarity across independently sampled responses & $K$ & \citet{Manakul::2023aa} \\
Verbalized confidence & Black & Model-elicited (self-reported) confidence score & $1$ ($+$prompt) & \citet{Kadavath::2022aa,Tian::2023aa,Xiong::2024aa} \\
Ensemble/Bayesian & White/Grey & Disagreement across members, seeds, or prompts; decomposed AU/EU & $M$ & \citet{Kirsch::2025aa,Jayasekera::2025aa} \\
Conformal/selective & Calibration set & Coverage-calibrated output sets or abstention & wrapper & \citet{Ren::2023aa,Badshah::2026aa,Kang::2025aa} \\
\bottomrule
\end{tabularx}
\begin{tablenotes}\footnotesize
\item[a] White = requires logits/internal states; Black = output text only; Grey = requires multiple stochastic passes.
\item[b] Black-box if an external entailment/clustering model supplies the semantic grouping.
\end{tablenotes}
\end{threeparttable}
\end{table*}

\subsection{Likelihood- and consistency-based measures}
The simplest signals read off the token distribution model.
That is, the (length-normalized) sequence log-likelihood or per-token entropy serves as a confidence proxy, but it is length-biased and, as in classification (\autoref{sec:measures}), cannot by itself separate aleatoric from epistemic uncertainty.
Semantic approaches lift the ensemble-disagreement view of \autoref{sec:measures} from a fixed label set to a space of meanings.
\citet{Kuhn::2023aa} sample multiple generations, cluster them by bidirectional entailment (two answers share a meaning if and only if each entails the other under a natural language inference model), and compute semantic entropy over the clusters.
This scales to reliable hallucination detection~\citep{Farquhar::2024aa}. 
Black-box variants quantify uncertainty purely from sampled-response consistency, such as SelfCheckGPT~\citep{Manakul::2023aa}.

\subsection{Verbalized and elicited confidence}
A distinctively \gls{llm} capability is asking the model to state its confidence. \citet{Kadavath::2022aa} showed that models can be trained to predict whether their own answers are correct (``knowing what they know''), and prompting strategies elicit confidence scores whose calibration varies widely~\citep{Tian::2023aa,Xiong::2024aa}. 
These self-reports are cheap but fragile, and their calibration degrades under distribution shift and adversarial prompting.

\subsection{Bayesian, ensemble, and decomposition views}
The Bayesian, ensemble, and decomposition views developed above (\autoref{sec:measures}) transfer only partially to the generative setting.
Epistemic uncertainty can be captured through (implicit) ensembles, but \citet{Kirsch::2025aa} show that scale induces an epistemic-uncertainty collapse across nominally independent members (\autoref{sec:efficient}). 
The token- and sequence-level entropy and \gls{mi} decomposition was first extended from a fixed label set to autoregressive structured prediction using ensembles of autoregressive models~\citep{Malinin::2021aa}.
Decomposition ideas have since been adapted to the generative setting for natural-language generation and for in-context learning~\citep{Jayasekera::2025aa}, and uncertainty has been argued to be decision-critical for \gls{llm} agents~\citep{Felicioni::2024aa}.

\subsection{Downstream use: Selective generation and conformal methods}
As in \autoref{sec:ood}, the measure should support decision-making. 
Conformal and selective-generation methods provide coverage-style guarantees for conditional language models (those that generate an output conditioned on an input, as in summarization or translation)~\citep{Ren::2023aa} and for \gls{llm}-as-judge pipelines~\citep{Badshah::2026aa}, while self-certainty scores guide Best-of-$N$ response selection~\citep{Kang::2025aa}. 
Here ``coverage'' is the generative analogue of the classification guarantee. 
Rather than a prediction set that contains the true label with probability $1-\alpha$ (\textit{e.g.}, $90\%$ when $\alpha=0.1$), the method returns a set of sampled generations (or an abstention) that contains an acceptable response at that same level, under the caveats on exchangeability noted in \autoref{sec:conformal}.
As in the classification setting of \autoref{sec:measures}, most current scores are heuristic.
They are validated by correlation with downstream errors rather than derived as estimators of a defined quantity. 
Semantic entropy~\citep{Kuhn::2023aa}, response-consistency checks~\citep{Manakul::2023aa}, and verbalized confidence~\citep{Tian::2023aa} each supply a usable signal, but none targets a posterior, a proper score, or a coverage level, and so none carries a guarantee that survives a change of model, prompt, or domain. 
A principled measure, by contrast, would estimate a well-specified object and retain its calibration or coverage property under those shifts; conformal generation~\citep{Ren::2023aa} is the clearest step in this direction but is still confined to narrow output settings; although, non-exchangeable variants relax the exchangeability assumption via nearest-neighbor calibration~\citep{Ulmer::2024aa}.
Building measures that are principled, calibration-stable, and semantically aware, remains the central open problem in this area.

\section{Summary and open research directions}
\label{sec:summary}
The organizing contribution of this survey is a separation we have applied throughout. 
The method that produces a predictive ensemble is distinct from the measure that summarizes its uncertainty and the measures to evaluate the quality of the uncertainty estimates (\autoref{fig:roadmap}).
The two stages communicate only through the set of member predictions, so in principle any method can be paired with any measure, yet the literature has largely studied them apart.
This interchangeability is strongest among methods that expose an explicit set of member predictions (the explicit-member and posterior-sampling rows of \autoref{tab:scope}), for which the sample-based measures (predictive entropy, the mutual-information decomposition, and the pairwise divergences) are freely substitutable; closed-form single-pass methods return only a marginal predictive plus a method-specific epistemic signal (Dirichlet precision, feature density, or GP variance), so the sample-based decompositions apply to them only once their predictive is sampled, which moves them up a row in \autoref{tab:scope}.
Reading the field through this lens clarifies that progress on generating diverse predictions does not automatically transfer to reading uncertainty off them, and that comparisons are meaningful only against a stated evidentiary basis (\autoref{sec:eval}).

\subsection{Key findings}
The survey is consolidated in two sets of comparison tables placed with their subject matter: 
(i) the method tables (\autoref{tab:methods} and \autoref{tab:methods-procon}, in \autoref{sec:method-summary}; the taxonomy of \autoref{tab:scope} in \autoref{sec:framework}); and
(ii) the measure tables (\autoref{tab:measures}--\autoref{tab:measure-class}, in \autoref{sec:measures}), their qualitative ratings should be read against \autoref{sec:eval}.
From these, several patterns emerge.
First, there is a consistent trade-off between computational cost and uncertainty quality.
Methods that explore multiple posterior modes (deep ensembles, \gls{cde}, \gls{mswag}) produce superior uncertainty but at the cost of training multiple models.
Efficient approximations via weight sharing, trajectory collection, or stochastic masking invariably sacrifice diversity and yield less informative uncertainty under shift.
Second, closed-form predictive methods (those that read uncertainty analytically from a single deterministic pass) can nonetheless achieve strong uncertainty with appropriate inductive biases: \gls{sngp}, \gls{ddu}, and prior/posterior networks pursue this by different means.
Third, the classification in \autoref{tab:scope} explains why deep ensembles and their extensions lead benchmarks; they are the explicit-member methods that operate over the full network, the only cell in which between-mode (multi-modal) diversity is achieved.
Fourth, for measures, information-theoretic estimators exist for all three uncertainty types in classification, whereas moment- and divergence-based alternatives are either regression-only or quadratic in $M$; efficient, non-entropy-based epistemic measures for classification remain under-explored.

\subsection{Open research questions}
\paragraph{Last-layer diversity and its limits} Whether last-layer diversity alone can recover most of the benefit of full deep ensembles impacts directly on efficient architecture design.
The extent to which shared gradients collapse classification heads, and what mitigates it, is open~\citep{Kirsch::2025aa}.

\paragraph{Efficient epistemic measures for classification} The demonstration that additive entropy decompositions can fail under finite ensembles and posterior mismatch \citep{Schweighofer::2023aa,Wimmer::2023aa} motivates measures that bypass entropy identities at sub-quadratic cost.
A concrete near-term step is to validate bounded pairwise measures (\gls{epjs}) against \gls{mi} and \gls{epkl} on the \gls{ood}/shift suites of \autoref{sec:eval}, closing the gap between the structural argument of \autoref{sec:stability} and empirical performance.

\paragraph{Diversity and calibration under shift} The relationship between ensemble diversity and calibration under covariate shift needs deeper theory; current diversity metrics are largely empirical.

\paragraph{Hybrid architectures} Since a method that produces predictions and a measure that summarizes them are separable (\autoref{fig:roadmap}), single-pass techniques can in principle be combined with ensembling.
For example, a small deep ensemble of \gls{sngp} or \gls{ddu} backbones, or a single distance-aware backbone carrying several last-layer heads, would add between-mode diversity to a distance-aware single-pass model at a fraction of cost of full ensembles. 
Whether such hybrids recover most of the uncertainty quality of a deep ensemble for that added cost is an open question; systematic evaluation is lacking.

\paragraph{Input-dependent uncertainty structure} The classification analogue of heteroscedastic regression (that ensemble reliability varies across inputs and classes) suggests per-class uncertainty structure that current aggregation largely ignores.

\paragraph{Explaining uncertainty estimates} Beyond producing and measuring uncertainty, explaining \emph{why} a model is uncertain about a given input (\autoref{sec:ood}) remains under-explored. Counterfactual approaches such as \gls{clue} are increasingly relevant for trust and deployment but remain largely absent from the ensemble-and-measure pipeline surveyed here.

\paragraph{Uncertainty in large language models} The most consequential open frontier is generative \gls{uq} (\autoref{sec:llm}). 
The ensemble-and-measure framework transfers only partially to this setting.
Semantic entropy~\citep{Kuhn::2023aa,Farquhar::2024aa} is the generative analogue of the predictive-entropy decomposition, and sampled-response consistency~\citep{Manakul::2023aa} mirrors the pairwise-divergence measures of \autoref{sec:pairwise}. 
Three questions stand out: (i) whether a principled aleatoric/epistemic split exists for free-form generation~\citep{Jayasekera::2025aa}; (ii) whether verbalized confidence can be made calibration-stable under distribution shift and adversarial prompting~\citep{Tian::2023aa,Xiong::2024aa,Kadavath::2022aa}; and (iii) whether the epistemic-collapse of large models~\citep{Kirsch::2025aa} can be counteracted without the prohibitive cost of full ensembles. 

\paragraph{Conclusion}
On the method side the evidence is consistent.
Full-network, multi-modal exploration, deep ensemble, and their multi-modal extensions continue to set the standard for uncertainty quality, while efficient single-pass and last-layer approaches recover much of it when equipped with the right inductive bias, but do not capture the between-mode epistemic component.
On the measure side the picture is less settled.
The additive entropy/mutual-information decomposition remains the default despite known failures under finite ensembles and posterior mismatch, and bounded or efficient alternatives such as \gls{epjs} and variance-gated scores are structurally attractive but still await systematic empirical validation.
The practitioner's choice therefore remains a deliberate cost--quality trade-off rather than a single dominant framework.
The same framework indicates where the field is heading.
The most consequential frontier is generative \gls{uq} (\autoref{sec:llm}), which requires lifting the diversity, decomposition, and calibration tools surveyed here from a fixed label set into a space of semantic equivalences, where even the aleatoric/epistemic split must be redefined.
Whether the method--measure separation that organized this survey survives that move is, in our view, the most important question it leaves open.

\bibliographystyle{elsarticle-harv}
\bibliography{references}

\end{document}